\documentclass[12pt]{article}

\usepackage{scicite}

\usepackage{amssymb}
\usepackage{amsmath}
\usepackage{amsmath}
\usepackage{amsfonts}
\usepackage{amsthm}
\usepackage{multirow}
\usepackage{comment}
\usepackage[autostyle, english = american]{csquotes}
\MakeOuterQuote{"}

\usepackage{blindtext}
\usepackage{graphicx}
\usepackage{bbm}
\usepackage{hyperref}
\usepackage{dsfont}
\usepackage{fontawesome} 
\usepackage{calligra}

\usepackage{times}

\usepackage[color=red]{todonotes}

\usepackage{lineno}

\newcommand{\figref}[1]{Figure~\ref{#1}}

\newcommand{\eqnref}[1]{Equation~\ref{#1}}
\def\eg{\emph{e.g.}}
\def\ie{\emph{i.e.}}

\definecolor{Gray}{gray}{0.95}

\newcommand{\prob}[1]{\mathbb{P}\left( #1 \right)}
\newcommand{\qprob}[1]{\mathbb{Q}\left( #1 \right)}

\def\vec#1{\mathchoice{\mbox{\boldmath $\displaystyle\bf#1$}}
{\mbox{\boldmath $\textstyle\bf#1$}}
{\mbox{\boldmath $\scriptstyle\bf#1$}}
{\mbox{\boldmath $\scriptscriptstyle\bf#1$}}}

\usepackage{amsmath,amsthm,amssymb}

\newcommand\cut[1]{}

\newcommand{\squishlist}{
   \begin{list}{$\bullet$}
    { \setlength{\itemsep}{0pt}      \setlength{\parsep}{3pt}
      \setlength{\topsep}{3pt}       \setlength{\partopsep}{0pt}
      \setlength{\leftmargin}{1.5em} \setlength{\labelwidth}{1em}
      \setlength{\labelsep}{0.5em} } }

\newcommand{\squishlisttwo}{
   \begin{list}{$\bullet$}
    { \setlength{\itemsep}{0pt}    \setlength{\parsep}{0pt}
      \setlength{\topsep}{0pt}     \setlength{\partopsep}{0pt}
      \setlength{\leftmargin}{2em} \setlength{\labelwidth}{1.5em}
      \setlength{\labelsep}{0.5em} } }

\newcommand{\squishend}{
    \end{list}  }

\newcommand{\myvec}[1]{\mathbf{#1}}

\newcommand{\vx}{\myvec{x}}

\newcommand{\E}{\mathbb{E}}

\newcommand{\be}{\begin{equation}}
\newcommand{\ee}{\end{equation}}
\newcommand{\bea}{\begin{eqnarray}}
\newcommand{\eea}{\end{eqnarray}}
\newcommand{\beaa}{\begin{eqnarray*}}
\newcommand{\eeaa}{\end{eqnarray*}}

\DeclareMathAlphabet{\mathpzc}{OT1}{pzc}{m}{n}

\newcommand*{\bdiv}{%
  \nonscript\mskip-\medmuskip\mkern5mu%
  \mathbin{\operator@font div}\penalty900\mkern5mu%
  \nonscript\mskip-\medmuskip
}

\newenvironment{sciabstract}{%
\begin{quote} \bf}
{\end{quote}}

\title{Human-level performance
in first-person multiplayer games
with population-based deep reinforcement learning}

\author
{\normalsize{Max Jaderberg$^{*1}$, Wojciech M. Czarnecki$^{*1}$, Iain Dunning$^{*1}$, Luke Marris$^{1}$}\\
\normalsize{Guy Lever$^1$, Antonio Garcia Castaneda$^1$, Charles Beattie$^1$, Neil C. Rabinowitz$^{1}$}\\
\normalsize{Ari S. Morcos$^{1}$, Avraham Ruderman$^{1}$, Nicolas Sonnerat$^1$, Tim Green$^1$, Louise Deason$^1$}\\
\normalsize{Joel Z. Leibo$^1$, David Silver$^1$, Demis Hassabis$^1$, Koray Kavukcuoglu$^1$, Thore Graepel$^1$}
\\
\small{$^*$Equal contribution.}
\\
\\
\normalsize{$^{1}$DeepMind, London, UK}\\
}

\date{}

\begin{document}

\maketitle

\begin{sciabstract}
Recent progress in artificial intelligence through reinforcement learning (RL) has shown great success on increasingly complex single-agent environments~\cite{MnihDQN, MnihA3C, SchulmanPPO, LillicrapDDPG, JaderbergUnreal} and two-player turn-based games~\cite{TesauroTDGammon,silver2017mastering,MoravcikDeepStack}.
However, the real-world contains multiple agents, each learning and acting independently to cooperate and compete with other agents, and environments reflecting this degree of complexity remain an open challenge.
In this work, we demonstrate for the first time that an agent can
achieve human-level in a popular 3D multiplayer first-person video game, Quake III Arena Capture the Flag~\cite{QuakeThree}, using only pixels and game points as input.
These results were achieved by a novel two-tier optimisation process in which a population of independent RL agents are trained concurrently from thousands of parallel matches with agents playing in teams together and against each other on randomly generated environments.
Each agent in the population learns its own internal reward signal to complement the sparse delayed reward from winning, and selects actions using a novel temporally hierarchical representation that enables the agent to reason at multiple timescales.
During game-play, these agents display human-like behaviours such as navigating, following, and defending based on a rich learned representation that is shown to encode high-level game knowledge.
In an extensive tournament-style evaluation the trained agents exceeded the win-rate of strong human players both as teammates and opponents, and proved far stronger than existing state-of-the-art agents.
These results demonstrate a significant jump in the capabilities of artificial agents, bringing us closer to the goal of human-level intelligence.
\end{sciabstract}

\begin{figure}[t]
    \centering
    \includegraphics[width=\textwidth]{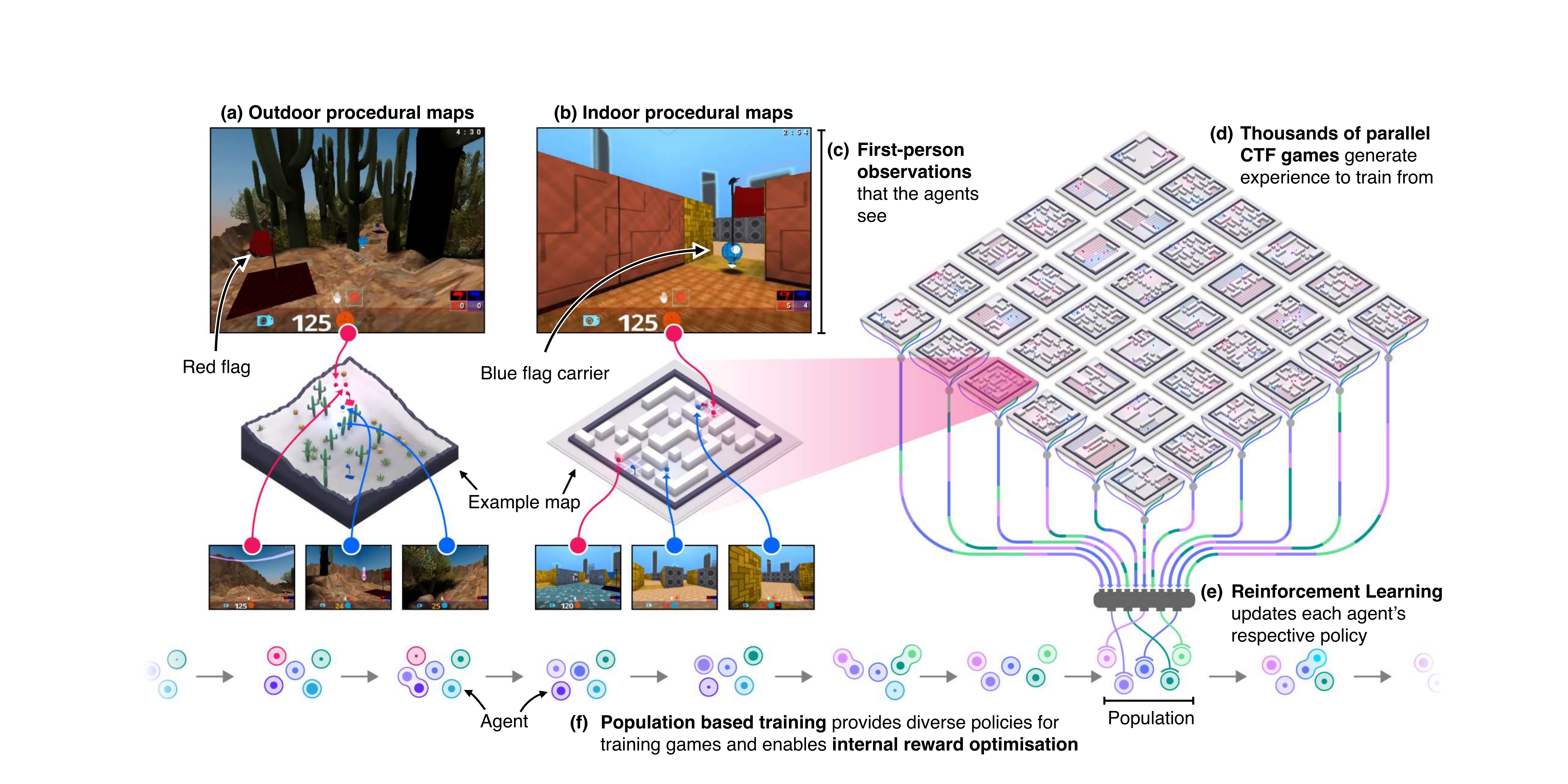}
        \caption{{\bf CTF task and computational training framework.} \small Shown are two example maps that have been sampled from the distribution of outdoor maps (a) and indoor maps (b). Each agent in the game only sees its own first-person pixel view of the environment (c). Training data is generated by playing thousands of CTF games in parallel on a diverse distribution of procedurally generated maps (d), and used to train the agents that played in each game with reinforcement learning (e). We train a population of 30 different agents together, which provides a diverse set of teammates and opponents to play with, and is also used to evolve the internal rewards and hyperparameters of agents and learning process (f). Game-play footage and further exposition of the environment variability can be found in Supplementary Video \url{https://youtu.be/dltN4MxV1RI}.}
    \label{fig:one}
\end{figure}

We demonstrate how intelligent behaviour can emerge from training sophisticated new learning agents within complex multi-agent environments.
End-to-end reinforcement learning methods~\cite{MnihDQN, MnihA3C} have so far not succeeded in training agents in multi-agent games that combine team and competitive play due to the high complexity of the learning problem~\cite{BernsteinDecPomdp,MatignonIndependentLearners} that arises from the concurrent adaptation of other learning agents in the environment.
We approach this challenge by studying team-based multiplayer 3D first-person video games, a genre which is particularly immersive for humans~\cite{ermi2005fundamental} and has even been shown to improve a wide range of cognitive abilities~\cite{green2015action}.
We focus specifically on a modified version~\cite{beattie2016deepmind} of Quake III Arena~\cite{QuakeThree}, the canonical multiplayer 3D first-person video game, whose game mechanics served as the basis for many subsequent games, and which has a thriving professional scene~\cite{QuakeCon}.
The task we consider is the game mode Capture the Flag (CTF) on per game randomly generated maps of both indoor and outdoor theme (\figref{fig:one} (a,b)).
Two opposing teams consisting of multiple individual players compete to capture each other's flags by strategically navigating, tagging, and evading opponents. The team with the greatest number of flag captures after five minutes wins.
CTF is played in a visually rich simulated physical environment (Supplementary Video \url{https://youtu.be/dltN4MxV1RI}), and agents interact with the environment and with other agents through their actions and observations.
In contrast to previous
work~\cite{silver2017mastering,MoravcikDeepStack,foerster2017learning,LoweMADDPG,mordatch2017emergence,NIPS2016_6398, riedmiller2007experiences,stone2000layered,LNAI17-MacAlpine2}, agents do not have access to models of the environment, other agents, or human policy priors, nor can they communicate with each other outside of the game environment. Each agent acts and learns independently, resulting in decentralised control within a team.

Since we wish to develop a learning agent capable of acquiring generalisable skills, we go beyond training fixed teams of agents on a fixed map, and instead devise an algorithm and training procedure that enables agents to acquire policies that are robust to the variability of maps, number of players, and choice of teammates, a paradigm closely related to ad-hoc team play~\cite{stone2010ad}.
The proposed training algorithm stabilises the learning process in partially observable multi-agent environments by concurrently training a diverse population of agents who learn by playing with each other, and in addition the agent population provides a mechanism for meta-optimisation.
We solve the prohibitively hard credit assignment problem of learning from the sparse and delayed episodic team win/loss signal (optimising thousands of actions based on a single final reward) by enabling agents to evolve an internal reward signal that acts as a proxy for winning and provides denser rewards.
Finally, we meet the memory and long-term temporal reasoning requirements of high-level, strategic CTF play by introducing an agent architecture that features a multi-timescale representation, reminiscent of what has been observed in primate cerebral cortex~\cite{chen2015processing}, and an external working memory module, broadly inspired by human episodic memory~\cite{hassabis2017neuroscience}.
These three innovations, integrated within a scalable, massively distributed, asynchronous computational framework, enables the training of highly skilled CTF agents through solely multi-agent interaction and single bits of feedback about game outcomes.

\begin{figure}[h!]
    \centering
    \includegraphics[width=\textwidth]{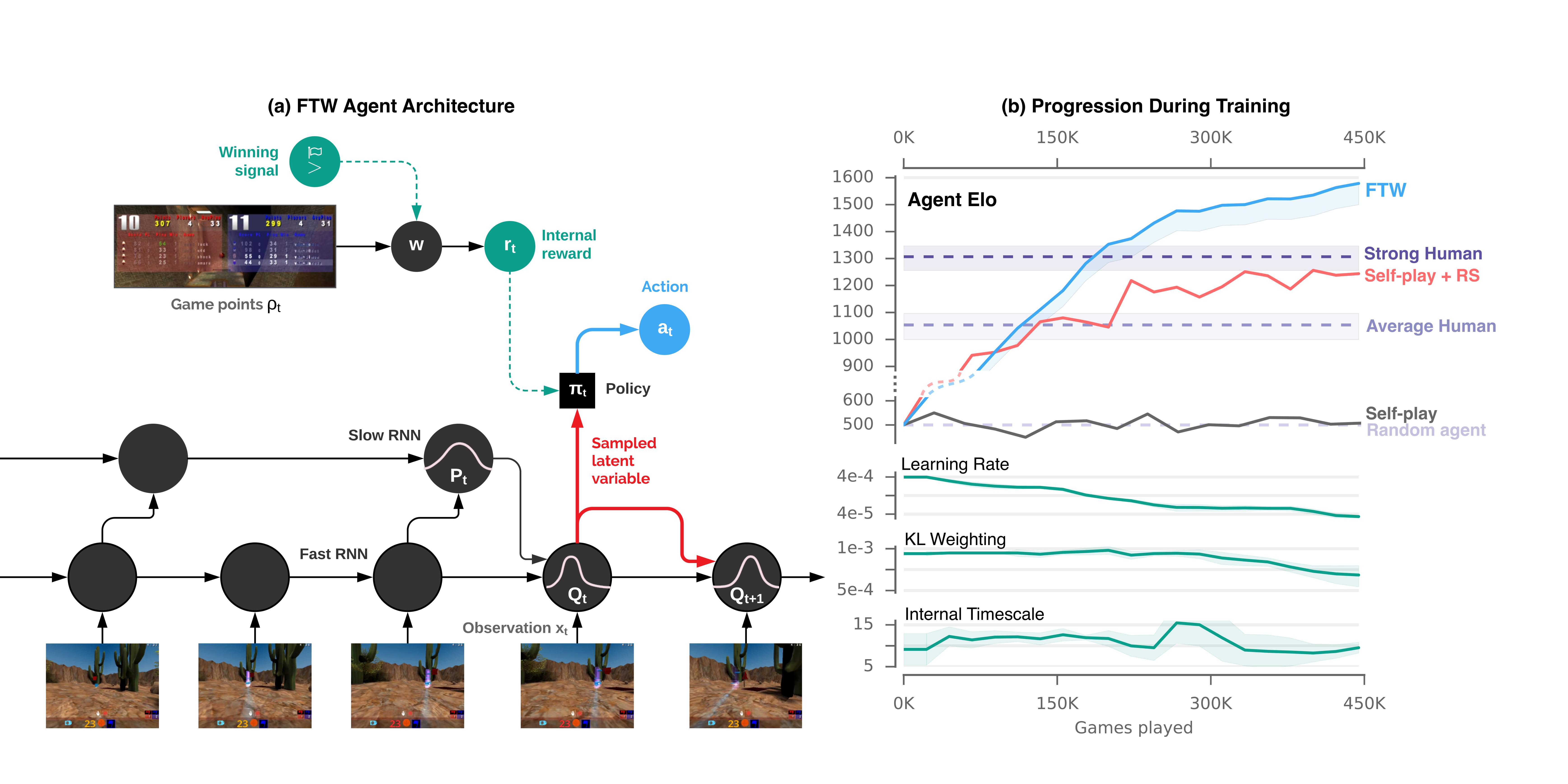}
    \caption{{\bf Agent architecture and benchmarking.} \small
    (a) Shown is how the agent processes a temporal sequence of observations $\vec{x}_t$ from the environment. The model operates at two different time scales, faster at the bottom, and slower by a factor of $\tau$ at the top. A stochastic vector-valued latent variable is sampled at the fast time scale from distribution $\mathbb{Q}_t$ based on observations $\vec{x}_t$. The action distribution $\pi_t$ is sampled conditional on the latent variable at each time step $t$. The latent variable is regularised by the slow moving prior $\mathbb{P}_t$ which helps capture long-range temporal correlations and promotes memory. The network parameters are updated using reinforcement learning based on the agent's own internal reward signal $r_t$, which is obtained from a learnt transformation $\vec{w}$ of game points $\rho_{t}$. $\vec{w}$ is optimised for winning probability through population based training, another level of training performed at yet a slower time scale than RL. Detailed network architectures are described in Figure~\ref{fig:arch}.
    (b) Top: Shown are the Elo skill ratings of the FTW agent population throughout training (blue) together with those of the best baseline agents using hand tuned reward shaping (RS) (red) and game winning reward signal only (black), compared to human and random agent reference points (violet, shaded region shows strength between 10th and 90th percentile). It can be seen that the FTW agent achieves a skill level considerably beyond strong human subjects, whereas the baseline agent's skill plateaus below, and does not learn anything without reward shaping (see Supplementary Materials for evaluation procedure).
    (b) Bottom: Shown is the evolution of three hyperparameters of the FTW agent population: learning rate, KL weighting, and internal time scale $\tau$, plotted as mean and standard deviation across the population.}
    \label{fig:two}
\end{figure}

In our formulation, the agent's policy $\pi$ uses the same interface available to human players. It receives raw RGB pixel input $\vec{x}_t$ from the agent's first-person perspective at timestep $t$, produces control actions $a_t\sim \pi$ simulating a gamepad, and receives game points $\rho_t$ attained -- the points received by the player for various game events which is visible on the in-game scoreboard.
The goal of reinforcement learning (RL) is to find a policy that maximises the expected cumulative $\gamma$-discounted reward $\E_{\pi}[\sum_{t=0}^T \gamma^{t} r_t]$ over a CTF game with $T$ time steps.
The agent's policy $\pi$ is parameterised by a multi-timescale recurrent neural network with external memory~\cite{graves2016hybrid} (\figref{fig:two} (a), \figref{fig:arch}).
Actions in this model are generated conditional on a stochastic latent variable, whose distribution is modulated by a more slowly evolving prior process. The variational objective function encodes a trade-off between maximising expected reward and consistency between the two timescales of inference (more details are given in Supplementary Materials Section \ref{sec:ftwagent}).
Whereas some previous hierarchical RL agents construct explicit hierarchical goals or skills~\cite{SuttonOptions,VezhnevetsFun, BaconOptionCritic}, this agent architecture is conceptually more closely related to work on building hierarchical temporal representations~\cite{clockwork,chung2016hierarchical,schmidhuber1992learning,el1996hierarchical} and recurrent latent variable models for sequential data~\cite{chung2015recurrent,fraccaro2016sequential}.
The resulting model constructs a temporally hierarchical representation space in a novel way to promote the use of memory (Figure~\ref{fig:ext_dnc}) and temporally coherent action sequences.

For ad-hoc teams, we postulate that an agent's policy $\pi_0$ should maximise the probability of winning for its team, $\{\pi_0, \pi_1, \ldots, \pi_{\frac{N}{2}-1}\}$, which is composed of $\pi_0$ itself, and its teammates' policies $\pi_1, \ldots, \pi_{\frac{N}{2}-1}$, for a total of $N$ players in the game:
\begin{equation}
    \mathbbm{P}(\text{$\pi_0$'s team wins}| \omega, ( \pi_n )_{n=0}^{N-1} ) =
    \E_{\vec{a} \sim ( \pi_n)_{n=0}^{N-1}}\left[
    \{ \pi_0, \pi_1, \ldots, \pi_{\frac{N}{2}-1} \}
    \overset{\text{\faFlagO}}{>}
    \{ \pi_{\frac{N}{2}}, \ldots, \pi_{N-1}\}
    \right].
\label{eqn:winning}
\end{equation}
%
The winning operator $\overset{\text{\faFlagO}}{>}$ returns 1 if the left team wins, 0 for losing, and randomly breaks ties.
$\omega \sim \Omega$ represents the specific map instance and random seeds, which are stochastic in learning and testing.
Since game outcome as the only reward signal is too sparse for RL to be effective, we require rewards $r_t$ to direct the learning process towards winning yet are more frequently available than the game outcome.
In our approach, we operationalise the idea that each agent has a dense internal reward function~\cite{singh2009rewards,singh2010intrinsically,wolpert1999introduction}, by specifying $r_t = \vec{w}(\rho_t)$ based on the available game points signals $\rho_t$ (points are registered for events such as capturing a flag), and, crucially, allowing the agent to learn the transformation $\vec{w}$ such that policy optimisation on the internal rewards $r_t$ optimises the policy {\bf F}or {\bf T}he {\bf W}in, giving us the \emph{FTW agent}.

Training agents in multi-agent systems requires instantiations of other agents in the environment, like teammates and opponents, to generate learning experience.
A solution could be self-play RL, in which an agent is trained by playing against its own policy.
While self-play variants can prove effective in some multi-agent games~\cite{SilverAlphaGo,silver2017mastering,MoravcikDeepStack,BansalEmergentComplexity, BrownFictitiousPlay, LanctotPSRO, HeinrichDeepFictitious}, these methods can be unstable and in their basic form do not support concurrent training which is crucial for scalability.
Our solution is to train a population of $P$ different agents
$\vec{\pi} = (\pi_p)_{p=1}^{P}$ in parallel that play with each other, introducing diversity amongst players to stabilise training~\cite{rosin1997new}.
Each agent within this population learns from experience generated by playing with teammates and opponents sampled from the population.
We sample the agents indexed by $\iota$ for a training game using a stochastic matchmaking scheme $m_p(\vec{\pi})$ that biases co-players to be of similar skill to player $p$.
This scheme ensures that -- a priori -- the outcome is sufficiently uncertain to provide a meaningful learning signal, and that a diverse set of teammates and opponents are seen during training.
Agents' skill levels are estimated online by calculating Elo scores (adapted from chess~\cite{ELO}) based on outcomes of training games.
We also use the population to meta-optimise the internal rewards and hyperparameters of the RL process itself, which results in the joint maximisation of:
\begin{equation}
\begin{aligned}
J_\mathrm{inner}( \pi_p | \vec{w}_p) &= \mathbb{E}_{\iota \sim m_p(\vec{\pi}), \omega \sim \Omega}
\; \mathbb{E}_{\vec{a}\sim \vec{\pi}_{\iota}} \left [ \sum_{t=0}^T \gamma^t \vec{w}_p(\rho_{p,t}) \right ] \;\;\; \forall \pi_p \in \vec{\pi}\\
J_\mathrm{outer}( \vec{w}_p, \vec{\phi}_p | \vec{\pi} ) &= \mathbb{E}_{\iota \sim m_p(\vec{\pi}), \omega \sim \Omega}
\; \prob{\pi_p^{\vec{w},\vec{\phi}}\text{'s team wins}|\omega, \vec{\pi}^{\vec{w},\vec{\phi}}_{\iota}}
\end{aligned}
\end{equation}
$$
\pi^{\vec{w},\vec{\phi}}_p = \text{optimise}_{\pi_p}(J_\mathrm{inner}, \vec{w}, \vec{\phi}).
$$
This can be seen as a two-tier reinforcement learning problem.
The inner optimisation maximises $J_\mathrm{inner}$, the agents' expected future discounted internal rewards.
The outer optimisation of $J_\mathrm{outer}$ can be viewed as a meta-game, in which the meta-reward of winning the match is maximised with respect to internal reward schemes $\vec{w}_p$ and hyperparameters $\vec{\phi}_p$, with the inner optimisation providing the meta transition dynamics.
We solve the inner optimisation with RL as previously described, and the outer optimisation with Population Based Training (PBT)~\cite{jaderberg2017population}. PBT is an online evolutionary process which adapts internal rewards and hyperparameters and performs model selection by replacing under-performing agents with mutated versions of better agents.
This joint optimisation of the agent policy using RL together with the optimisation of the RL procedure itself towards a high-level goal proves to be an effective and generally applicable strategy, and utilises the potential of combining learning and evolution~\cite{ackley1991interactions} in large scale learning systems.

To assess the generalisation performance of agents at different points during training, we performed a large tournament on procedurally generated maps with ad-hoc matches involving three types of agents as teammates and opponents: ablated versions of FTW (including state-of-the-art baselines), Quake III Arena scripted bots of various levels~\cite{waveren2001quakebots}, and human participants with first-person video game experience.
\figref{fig:two}~(b) and Figure~\ref{tab:leaderboard_proc} show the Elo scores and derived winning probabilities for different ablations of FTW, and how the combination of components provide superior performance.
The FTW agents clearly exceeded the win-rate of humans in maps which neither agent nor human had seen previously, \ie~zero-shot generalisation, with a team of two humans on average capturing 16 flags per game less than a team of two FTW agents (Figure~\ref{tab:leaderboard_proc} Bottom, FF vs hh).
Interestingly, only as part of a human-agent team did we observe a human winning over an agent-agent team (5\% win probability).
This result suggests that trained agents are capable of cooperating with never seen before teammates, such as humans.
In a separate study, we probed the exploitability of the FTW agent by allowing a team of two professional games testers with full communication to play continuously against a fixed pair of FTW agents.
Even after twelve hours of practice the human game testers were only able to win 25\% (6.3\% draw rate) of games against the agent team.

Interpreting the difference in performance between agents and humans must take into account the subtle differences in observation resolution, frame rate, control fidelity, and intrinsic limitations in reaction time and sensorimotor skills (Figure \ref{fig:humanagentdiff} (a), Supplementary Materials Section~\ref{sec:humandiff}). For example, humans have superior observation and control resolution -- this may be responsible for humans successfully tagging at long range where agents could not (humans: 17\% tags above 5 map units, agents: 0.5\%). In contrast, at short range, agents have superior tagging reaction times to humans: by one measure FTW agents respond to newly appeared opponents in 258ms, compared with 559ms for humans (Figure~\ref{fig:humanagentdiff} (b)). Another advantage exhibited by agents is their tagging accuracy, where FTW agents achieve 80\% accuracy compared to humans' 48\%. By artificially reducing the FTW agents' tagging accuracy to be similar to humans (without retraining them), agents' win-rate was reduced, though still exceeded that of humans (Figure~\ref{fig:humanagentdiff} (c)). Thus, while agents learn to make use of their potential for better tagging accuracy, this is only one factor contributing to their overall performance.

\begin{figure}[h!]
    \centering
    \includegraphics[width=\textwidth]{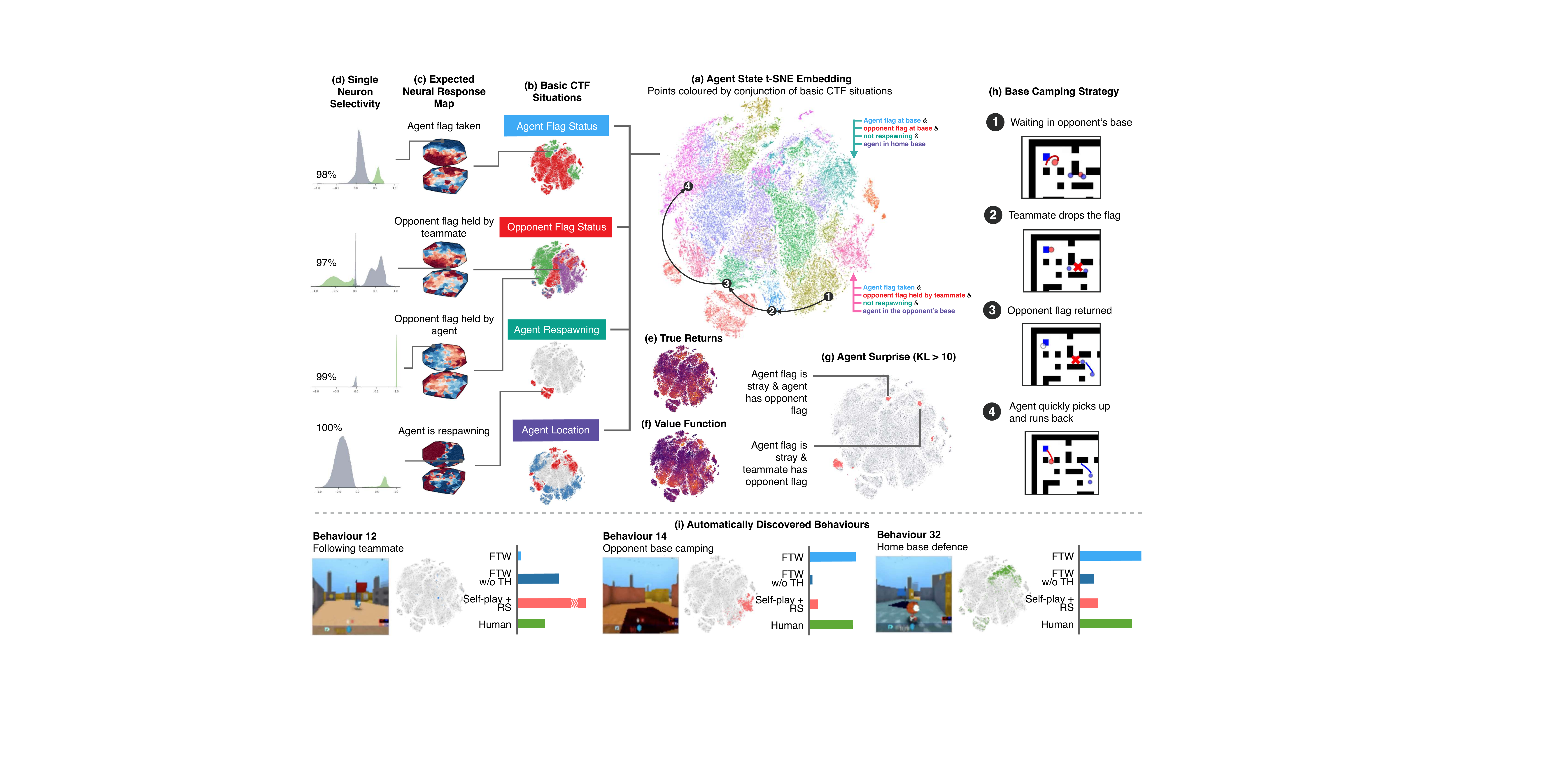}
    \caption{{\bf Knowledge representation and behavioural analysis.} \small
    (a)~The 2D t-SNE embedding of an FTW agent's internal states during game-play. Each point represents the internal state $(\vec{h}^p, \vec{h}^q)$ at a particular point in the game, and is coloured according to the high-level game state at this time -- the conjunction of four basic CTF situations (b). Colour clusters form, showing that nearby regions in the internal representation of the agent correspond to the same high-level game state. (c)~A visualisation of the expected internal state arranged in a similarity-preserving topological embedding (Figure~\ref{fig:ext_neural_response}). (d)~We show distributions of situation conditional activations for particular single neurons which are distinctly selective for these CTF situations, and show the predictive accuracy of this neuron. (e)~The true return of the agent's internal reward signal and (f) the agent's prediction, its value function. (g)~Regions where the agent's internal two-timescale representation diverges, the agent's surprise. (h) The four-step temporal sequence of the high-level strategy {\it opponent base camping}. (i)~Three automatically discovered high-level behaviours of agents and corresponding regions in the t-SNE embedding. To the right, average occurrence per game of each behaviour for the FTW agent, the FTW agent without temporal hierarchy (TH), self-play with reward shaping agent, and human subjects (more detail in Figure~\ref{fig:ext_behvaiours}).}
    \label{fig:three}
\end{figure}

We hypothesise that trained agents of such high skill have learned a rich representation of the game.
To investigate this, we extracted ground-truth state from the game engine at each point in time in terms of 200 binary features such as ``Do I have the flag?'', ``Did I see my teammate recently?'', and ``Will I be in the opponent's base soon?''.
We say that the agent has knowledge of a given feature if logistic regression on the internal state of the agent accurately models the feature.
In this sense, the internal representation of the agent was found to encode a wide variety of knowledge about the game situation (Figure \ref{fig:ext_knowledge}).
Interestingly, the FTW agent's representation was found to encode features related to the past particularly well: \eg~the FTW agent was able to classify the state \emph{both flags are stray} (flags dropped not at base) with 91\% AUCROC (area under the receiver operating characteristic curve), compared to 70\% with the self-play baseline.
Looking at the acquisition of knowledge as training progresses, the agent first learned about its own base, then about the opponent's base, and picking up the flag. Immediately useful flag knowledge was learned prior to knowledge related to tagging or one's teammate's situation.
Note that agents were never explicitly trained to model this knowledge, thus these results show the spontaneous emergence of these concepts purely through RL-based training.

A visualisation of how the agent represents knowledge was obtained by performing dimensionality reduction of the agent's activations using t-SNE~\cite{maaten2008visualizing}. As can be seen from \figref{fig:three}, internal agent state clustered in accordance with conjunctions of high-level game state features: flag status, respawn state, and room type. We also found individual neurons whose activations coded directly for some of these features, \eg~a neuron that was active if and only if the agent's teammate was holding the flag, reminiscent of concept cells~\cite{quiroga2012concept}. This knowledge was acquired in a distributed manner early in training (after 45K games), but then represented by a single, highly discriminative neuron later in training (at around 200K games).
This observed disentangling of game state is most pronounced in the FTW agent (Figure \ref{fig:ext_tsnes}).

\begin{figure}[h!]
    \centering
    \includegraphics[width=\textwidth]{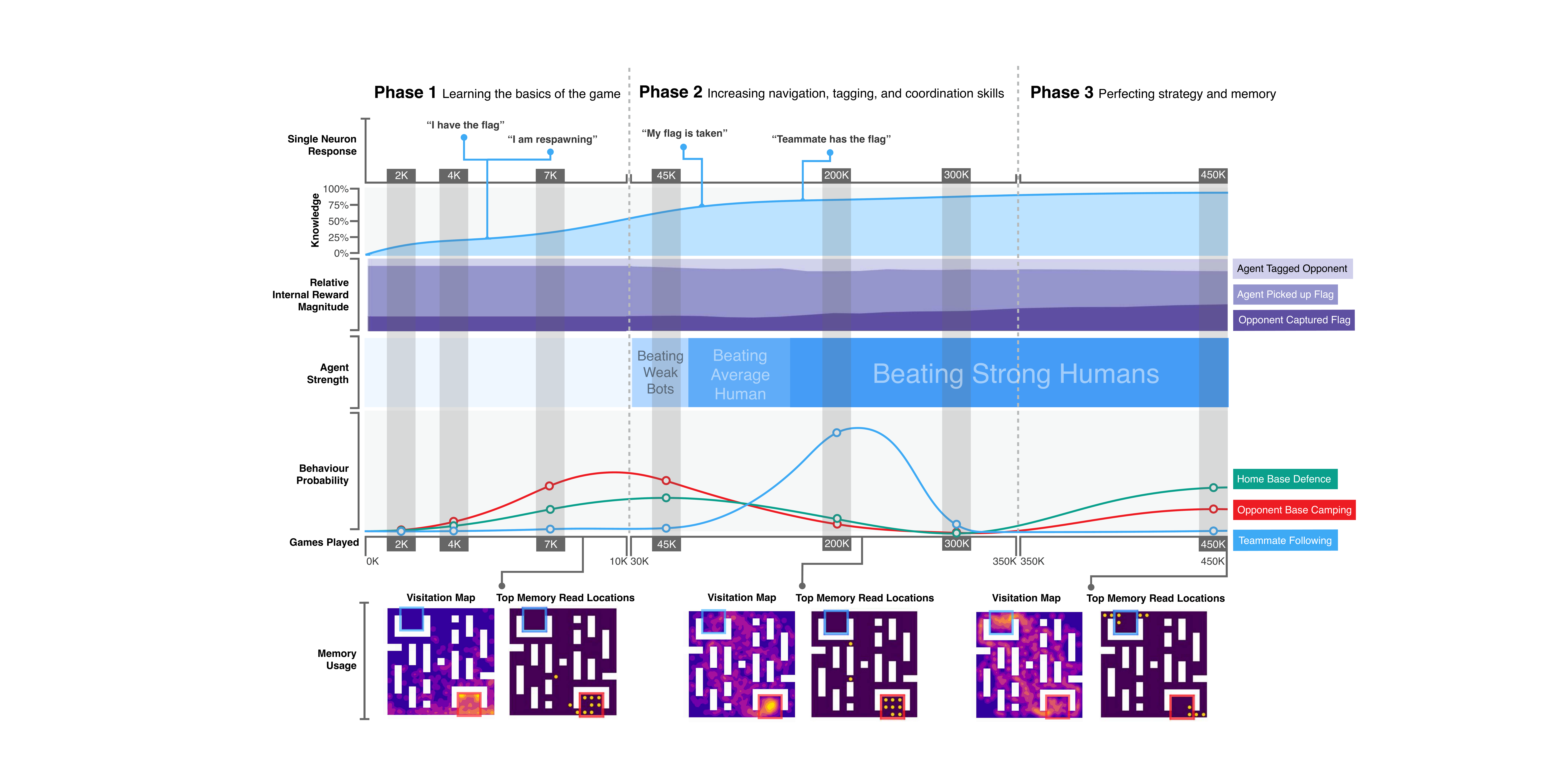}
    \caption{{\bf Progression of agent during training.} \small Shown is the development of knowledge representation and behaviours of the FTW agent over the training period of 450K games, segmented into three phases (Supplementary Video \url{https://youtu.be/dltN4MxV1RI}).
    {\bf Knowledge:} Shown is the percentage of game knowledge that is linearly decodable from the agent's representation, measured by average scaled AUCROC across 200 features of game state. Some knowledge is compressed to single neuron responses (Figure~\ref{fig:three} (a)), whose emergence in training is shown at the top.
    {\bf Relative Internal Reward Magnitude:} Shown is the relative magnitude of the agent's internal reward weights of three of the thirteen events corresponding to game points $\rho$. Early in training, the agent puts large reward weight on picking up the opponent flag, whereas later this weight is reduced, and reward for tagging an opponent and penalty when opponents capture a flag are increased by a factor of two.
    {\bf Behaviour Probability:} Shown are the frequencies of occurrence for three of the 32 automatically discovered behaviour clusters through training. {\it Opponent base camping} (red) is discovered early on, whereas {\it teammate following} (blue) becomes very prominent midway through training before mostly disappearing. The {\it home base defence} behaviour (green) resurges in occurrence towards the end of training, in line with the agent's increased internal penalty for more opponent flag captures.
    {\bf Memory Usage:} Shown are heat maps of visitation frequencies for locations in a particular map (left), and locations of the agent at which the top-ten most frequently read memories were written to memory, normalised by random reads from memory, indicating which locations the agent \emph{learned} to recall.
    Recalled locations change considerably throughout training, eventually showing the agent recalling the entrances to both bases, presumably in order to perform more efficient navigation in unseen maps, shown more generally in Figure \ref{fig:ext_dnc}.
    }
    \label{fig:four}
\end{figure}

One of the most salient aspects of the CTF task is that each game takes place on a randomly generated map, with walls, bases, and flags in new locations. We hypothesise that this requires agents to develop rich representations of these spatial environments to deal with task demands, and that the temporal hierarchy and explicit memory module of the FTW agent help towards this. An analysis of the memory recall patterns of the FTW agent playing in indoor environments shows precisely that: once the agent had discovered the entrances to the two bases, it primarily recalled memories formed at these base entrances (\figref{fig:four}, Figure~\ref{fig:ext_dnc}). We also found that the full FTW agent with temporal hierarchy learned a coordination strategy during maze navigation that ablated versions of the agent did not, resulting in more efficient flag capturing (Figure~\ref{tab:leaderboard_fetch_2vs2}).

Analysis of temporally extended behaviours provided another view on the complexity of behavioural strategies learned by the agent~\cite{krakauer2017neuroscience}.
We developed an unsupervised method to automatically discover and quantitatively characterise temporally extended behaviour patterns, inspired by models of mouse behaviour~\cite{wiltschko2015mapping}, which groups short game-play sequences into behavioural clusters (Figure~\ref{fig:ext_behvaiours}, Supplementary Video \url{https://youtu.be/dltN4MxV1RI}).
The discovered behaviours included well known tactics observed in human play, such as {\it waiting in the opponents base for a flag to reappear (opponent base camping)} which we only observed in FTW agents with a temporal hierarchy.
Some behaviours, such as {\it following a flag-carrying teammate}, were discovered and discarded midway through training, while others such as {\it performing home base defence} are most prominent later in training (\figref{fig:four}).

In this work, we have demonstrated that an artificial agent using only pixels and game points as input can learn to play highly competitively in a rich multi-agent environment: a popular multiplayer first-person video game.
This was achieved by combining a number of innovations in agent training -- population based training of agents, internal reward optimisation, and temporally hierarchical RL -- together with scalable computational architectures.
The presented framework of training populations of agents, each with their own learnt rewards, makes minimal assumptions about the game structure, and therefore should be applicable for scalable and stable learning in a wide variety of multi-agent systems, and the temporally hierarchical agent represents a sophisticated new architecture for problems requiring memory and temporally extended inference.
Limitations of the current framework, which should be addressed in future work, include the difficulty of maintaining diversity in agent populations, the greedy nature of the meta-optimisation performed by PBT, and the variance from temporal credit assignment in the proposed RL updates.
Trained agents exceeded the win-rate of humans in tournaments, and were shown to be robust to previously unseen teammates, opponents, maps, and numbers of players, and to exhibit complex and cooperative behaviours.
We discovered a highly compressed representation of important underlying game state in the trained agents, which enabled them to execute complex behavioural motifs.
In summary, our work introduces novel techniques to train agents which can achieve human-level performance at previously insurmountable tasks. When trained in a sufficiently rich multi-agent world, complex and surprising high-level intelligent artificial behaviour emerged.

\bibliographystyle{plain}

\section*{Acknowledgments}
We thank Matt Botvinick, Simon Osindero, Volodymyr Mnih, Alex Graves, Nando de Freitas, Nicolas Heess, and Karl Tuyls for helpful comments on the manuscript; Simon Green and Drew Purves for additional environment support and design; Kevin McKee and Tina Zhu for human experiment assistance; Amir Sadik and Sarah York for exploitation study participation; Adam Cain for help with figure design; Paul Lewis, Doug Fritz, and Jaume Sanchez Elias for 3D map visualisation work; Vicky Holgate, Adrian Bolton, Chloe Hillier, and Helen King for organisational support; and the rest of the DeepMind team for their invaluable support and ideas.

\renewcommand{\figurename}{Figure}
\setcounter{figure}{0}
\renewcommand{\thefigure}{S\arabic{figure}}

\renewcommand{\tablename}{Table}
\setcounter{table}{0}
\renewcommand{\thetable}{S\arabic{table}}

\section*{Supplementary Materials}

\section{Task}
\subsection{Rules of Capture the Flag}
CTF is a team game with the objective of scoring more flag captures than the opposing team in five minutes of play time. To score a \emph{capture}, a player must navigate to the opposing team's base, \emph{pick up} the flag (by touching the flag), \emph{carry} it back to their own base, and capture it by running into their own flag. A capture is only possible if the flag of the scoring player's team is safe at their base. Players may \emph{tag} opponents which teleports them back to their base after a delay (\emph{respawn}). If a \emph{flag carrier} is tagged, the flag they are carrying drops on the ground and becomes \emph{stray}. If a player on the team that owns the dropped flag touches the dropped flag, it is immediately returned back to their own base. If a player on the opposing team touches the dropped flag, that player will pick up the flag and can continue to attempt to capture the flag.

\subsection{Environment}
The environment we use is DeepMind Lab~\cite{beattie2016deepmind} which is a modified version of Quake III Arena~\cite{QuakeThree}. The modifications reduce visual connotations of violence, but retain all core game mechanics.
Video games form an important domain for research~\cite{laird2001human}. Previous work on first-person games considers either much simpler games~\cite{MnihA3C, JaderbergUnreal,wu2016training,lample2017playing}, simplified agent interfaces~\cite{van2009hierarchical}, or non-learning systems~\cite{orkin2006three,waveren2001quakebots}, and previously studied multi-agent domains often consist of discrete-state environments~\cite{leibo2017multi,NIPS2016_6398,foerster2017learning}, have simplified 2D dynamics~\cite{riedmiller2007experiences,LoweMADDPG,hausknecht2015deep} or have fully observable or non-perceptual features~\cite{LoweMADDPG,mordatch2017emergence,NIPS2016_6398,foerster2017learning,riedmiller2007experiences,hausknecht2015deep} rather than pixel observations. As an example, the RoboCup simulation league~\cite{kitano1997robocup} is a multi-agent environment that shares some of the same challenges of our environment, and successful work has included RL components~\cite{stone2000layered,LNAI17-MacAlpine2,riedmiller2007experiences}, however these solutions use a combination of hand-engineering, human-specified task decompositions, centralised control, and low-dimensional non-visual inputs, compared to our approach of end-to-end machine learning of independent reinforcement learners.

\begin{figure}[t]
    \centering
    \includegraphics[width=0.9\textwidth]{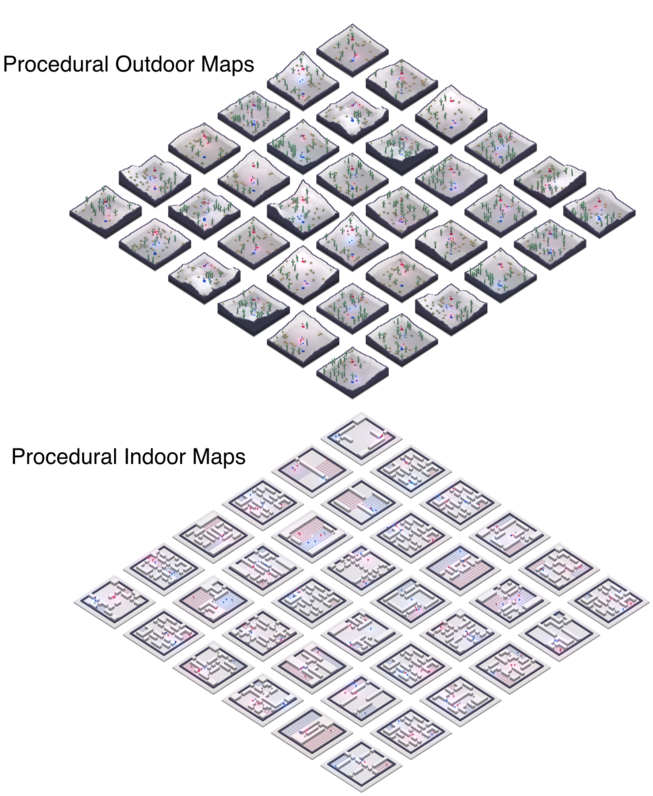}
    \caption{\small Shown are schematics of samples of procedurally generated maps on which agents were trained. In order to demonstrate the robustness of our approach we trained agents on two distinct styles of maps, procedural outdoor maps (top) and procedural indoor maps (bottom).}
    \label{fig:ext_maps}
\end{figure}

CTF games are played in an artificial environment referred to as a \emph{map}. In this work we consider two themes of procedurally generated maps in which agents play, indoor maps, and outdoor maps, example schematics of which are shown in Figure \ref{fig:ext_maps}. The \emph{procedural indoor maps} are flat, maze-like maps, rotationally symmetric and contain rooms connected by corridors. For each team there is a base room that contains their flag and player spawn points. Maps are contextually coloured: the red base is coloured red, the blue base blue. The \emph{procedural outdoor maps} are open and hilly naturalistic maps containing randomly sized rocks, cacti, bushes, and rugged terrain that may be impassable. Each team's flag and starting positions are located in opposite quadrants of the map. Both the procedural indoor maps and the procedural outdoor maps are randomly generated each episode (some random seeds are not used for training and reserved for performance evaluation), providing a very large set of environments. More details can be found in Section~\ref{sec:procmap}. Every player carries a disc gadget (equivalent to the railgun in Quake III Arena) which can be used for tagging, and
can see their team, shield, and flag status on screen.

\section{Agent}
\subsection{FTW Agent Architecture}\label{sec:ftwagent}
The agent's policy $\pi$ is represented by a neural network and optimised with reinforcement learning (RL).
In a fully observed Markov Decision Process, one would aim at finding a policy that maximises expected $\gamma$-discounted return $\E_{\pi(\cdot|\vec{s}_t)} [R_t]$ in game state $\vec{s}_t$, where $R_t = \sum_{k=0}^{T-t} \gamma^{k} r_{t+k}$.
However,
when an agent does not have information about the entire environment (which is often the case in real world problems, including CTF),
it becomes a Partially-Observed Markov Decision Process,
and hence we instead seek to maximise $\E_{\pi(\cdot|\vec{x}_{\leq t})} [R_t]$, the expected return under a policy conditioned on the agent's history of individual observations.
Due to the ambiguity of the true state given the observations, $\prob{\vec{s}_t | \vec{x}_{\leq t}}$, we represent the current value as a random variable,
$V_t = \E_{\pi(\cdot|\vec{x}_{\leq t})} [R_t] = \sum_{\vec{s}} \prob{\vec{s}|\mathbf{x}_{<t}}\E_{\pi(\cdot|\vec{s})} [R_t]$.
We follow the idea of RL as probabilistic inference~\cite{weber2015reinforced,levine2013variational,vlassis2009learning} which leads to a Kullback-Leibler divergence (KL) regularised objective in which the policy $\mathbb{Q}$ is regularised against a prior policy $\mathbb{P}$. We choose both to contain a latent variable $\vec{z}_t$, the purpose of which is to model the dependence on past observations. Letting the policy and the prior differ only in the way this dependence on past observations is modelled leads to the following objective:
\begin{equation}
   \E_{\qprob{\vec{z}_{t} | \mathrm{C}_t^q}} \left [ R_t\right ] - D_\text{KL}[\qprob{\vec{z}_{t} | \mathrm{C}_t^q} || \prob{\vec{z}_{t}| \mathrm{C}_t^p} ],
   \label{eqn:rlcost}
\end{equation}
\noindent where $\prob{\vec{z}_{t}| \mathrm{C}_t^p}$ and $\qprob{\vec{z}_{t} | \mathrm{C}_t^q} $ are the prior and variational posterior distributions on $\vec{z}_t$ conditioned on different sets of variables $\mathrm{C}_t^p$ and $\mathrm{C}_t^q$ respectively, and $D_\text{KL}$ is the Kullback-Leibler divergence.
The sets of conditioning variables $\mathrm{C}_t^p$ and $\mathrm{C}_t^q$ determine the structure of the probabilistic model of the agent, and can be used to equip the model with representational priors.
In addition to optimising the return as in Equation~\ref{eqn:rlcost}, we can also optimise extra modelling targets which are conditional on the latent variable $\vec{z}_t$, such as the value function to be used as a baseline~\cite{MnihA3C}, and pixel control~\cite{JaderbergUnreal}, whose optimisation positively shapes the shared latent representation.
The conditioning variables $\mathrm{C}_t^q$ and $\mathrm{C}_t^p$ and the associated neural network structure are chosen so as to promote forward planning and the use of memory.
We use a hierarchical RNN consisting of two recurrent networks (LSTMs~\cite{hochreiter1997long}) operating at different timescales.
The hierarchical RNN's fast timescale core generates a hidden state $\vec{h}_t^q$ at every environment step $t$, whereas its slow timescale core produces an updated hidden state every $\tau$ steps $\vec{h}_t^p = \vec{h}^p_{\tau \lfloor \frac{t}{\tau} \rfloor}$.
We use the output of the fast ticking LSTM as the variational posterior, $\qprob{\mathbf{z}_{t} | \prob{\mathbf{z}_t}, \mathbf{z}_{<t}, \mathbf{x}_{\leq t}, a_{<t}, r_{<t}} = \mathcal{N}(\mu^q_t, \Sigma_t^q)$, where the mean $\mu^q_t$ and covariance $\Sigma_t^q = (\sigma_t^q \mathbf{I})^2$ of the normal distribution are parameterised by the linear transformation $(\mu^q_t, \log \sigma_t^q) = f_q(\vec{h}_t^q)$, and at each timestep take a sample of $\vec{z}_t \sim \mathcal{N}(\mu^q_t, \Sigma_t^q)$.
The slow timescale LSTM output is used for the prior of $\prob{\mathbf{z}_{t}| \mathbf{z}_{< \tau \lfloor \frac{t}{\tau} \rfloor }, \mathbf{x}_{\leq \tau \lfloor \frac{t}{\tau} \rfloor }, a_{< \tau \lfloor \frac{t}{\tau} \rfloor }, r_{< \tau \lfloor \frac{t}{\tau} \rfloor }} = \mathcal{N}(\mu^p_t, \Sigma_t^p)$ where  $\Sigma_t^p = (\sigma_t^p \mathbf{I})^2$ , $(\mu^p_t, \log \sigma_t^p) = f_p(\vec{h}_t^p)$ and $f_p$ is a linear transformation.
The fast timescale core takes as input the observation that has been encoded by a convolutional neural network (CNN), $\vec{u}_t = \text{CNN}(\vec{x}_t)$, the previous action $a_{t-1}$, previous reward $r_{t-1}$, as well as the prior distribution parameters $\mu^p_t$ and $\Sigma_t^p$, and the previous sample of the variational posterior $\vec{z}_{t-1} \sim \mathcal{N}(\mu^q_{t-1}, \Sigma_{t-1}^q)$.
The slow core takes in the fast core's hidden state as input, giving the recurrent network dynamics of
\begin{linenomath}\begin{equation*}
    \vec{h}_t^q = g_q(\vec{u}_t, a_{t-1}, r_{t-1}, \vec{h}_t^p, \vec{h}_{t-1}^q, \mu^p_t, \Sigma_t^p, \vec{z}_{t-1})
\end{equation*}\end{linenomath}
\begin{equation}
   \vec{h}^p_t = \begin{cases}
    g_p(\vec{h}^q_{t-1}, \vec{h}^p_{t-1})  & \mbox{if}~ t \bmod \tau = 0 \\
    \vec{h}^p_{\tau \lfloor \frac{t}{\tau} \rfloor} & \mbox{otherwise}\\
\end{cases}
   \label{eqn:hierrnn}
\end{equation}
where $g_q$ and $g_p$ are the fast and slow timescale LSTM cores respectively. Stochastic policy, value function, and pixel control signals are obtained from the samples $\vec{z}_t$ using further non-linear transformations.
The resulting update direction is therefore:
\begin{equation}
   \nabla \big( \E_{\mathbf{z}_t \sim \mathbb{Q}} \left[-\mathcal{L}(\vec{z}_t, \vec{x}_t)\right] - D_\text{KL}\big[\mathbb{Q}(\mathbf{z}_{t} | \underset{\mathrm{C}_t^q}{\underbrace{\prob{\mathbf{z}_t}, \mathbf{z}_{<t}, \mathbf{x}_{\leq t}, a_{<t}, r_{<t}}}) || \mathbb{P}(\mathbf{z}_{t}| \underset{\mathrm{C}_t^p}{\underbrace{\mathbf{z}_{< \tau \lfloor \frac{t}{\tau} \rfloor }, \mathbf{x}_{\leq \tau \lfloor \frac{t}{\tau} \rfloor }, a_{< \tau \lfloor \frac{t}{\tau} \rfloor }, r_{< \tau \lfloor \frac{t}{\tau} \rfloor }}})\big] \big).
   \label{eqn:hiervar}
\end{equation}
where $\mathcal{L}(\cdot, \cdot)$ represents the objective function composed of terms for multi-step policy gradient and value function optimisation~\cite{MnihA3C}, as well as pixel control and reward prediction auxiliary tasks~\cite{JaderbergUnreal}, see Section~\ref{sec:supp_train}.
Intuitively, this objective function captures the idea that the slow LSTM generates a prior on $\vec{z}$ which predicts the evolution of $\vec{z}$ for the subsequent $\tau$ steps, while the fast LSTM generates a variational posterior on $\vec{z}$ that incorporates new observations, but adheres to the predictions made by the prior.
All the while, $\vec{z}$ must be a useful representation for maximising reward and auxiliary task performance.
This architecture can be easily extended to more than two hierarchical layers, but we found in practice that more layers made little difference on this task.
We also augmented this dual-LSTM agent with shared DNC memory~\cite{graves2016hybrid} to further increase its ability to store and recall past experience (this merely modifies the functional form of $g_p$ and $g_q$).
Finally, unlike previous work on DeepMind Lab~\cite{JaderbergUnreal, espeholt2018impala}, the FTW agent uses a rich action space of 540 individual actions which are obtained by combining elements from six independent action dimensions. Exact agent architectures are described in Figure~\ref{fig:arch}.

\subsection{Internal Reward and Population Based Training}
We wish to optimise the FTW agent with RL as stated in \eqnref{eqn:hiervar}, using a reward signal that maximises the agent team's win probability.
Reward purely based on game outcome, such as win/draw/loss signal as a reward of $r_T=1$, $r_T=0$, and $r_T=-1$ respectively, is very sparse and delayed, resulting in no learning (Figure~\ref{fig:two} (b) Self-play). Hence, we obtain more frequent rewards by considering the game points stream $\rho_t$. These can be used simply for reward shaping~\cite{ng1999policy} (Figure~\ref{fig:two} (b) Self-play + RS) or transformed into a reward signal $r_t = \vec{w}(\rho_t)$ using a learnt transformation $\vec{w}$ (Figure~\ref{fig:two} (b) FTW). This transformation is adapted such that performing RL to optimise the resulting cumulative sum of expected future discounted rewards effectively maximises the winning probability of the agent's team, removing the need for manual reward shaping~\cite{ng1999policy}.
The transformation $\vec{w}$ is implemented as a table look-up for each of the 13 unique values of $\rho_t$, corresponding to the events listed in Section~\ref{sec:events}.
In addition to optimising the internal rewards of the RL optimisation, we also optimise hyperparameters $\vec{\phi}$ of the agent and RL training process automatically. These include learning rate, slow LSTM time scale $\tau$, the weight of the $D_\text{KL}$ term in \eqnref{eqn:hiervar}, and the entropy cost (full list in Section~\ref{sec:supp_train}).
This optimisation of internal rewards and hyperparameters is performed using a process of population based training (PBT)~\cite{jaderberg2017population}.
In our case, a population of $P=30$ agents was trained in parallel. For each agent we periodically sampled another agent, and estimated the win probability of a team composed of only the first agent versus a team composed of only the second from training matches using Elo scores. If the estimated win probability of an agent was found to be less than 70\% then the losing agent copied the policy, the internal reward transformation, and hyperparameters of the better agent, and explored new internal rewards and hyperparameters. This exploration was performed by perturbing the inherited value by $\pm20$\% with a probability of 5\%, with the exception of the slow LSTM time scale $\tau$, which was uniformly sampled from the integer range $[5, 20)$.
A burn-in time of 1K games was used after each exploration step which prevents further exploration and allows learning to occur.

\subsection{Training Architecture}
We used a distributed, population-based training framework for deep reinforcement learning agents designed for the fast optimisation of RL agents interacting with each other in an environment with high computational simulation costs.
Our architecture is based on an actor-learner structure~\cite{espeholt2018impala}: a large collection of 1920 \emph{arena} processes continually play CTF games with players sampled at the beginning of each episode from the live training population to fill the $N$ player positions of the game (Section~\ref{sec:traingames} for details). We train with $N=4$ (2 vs 2 games) but find the agents generalise to different team sizes (Figure~\ref{tab:leaderboard_proc_nsvsn}).
After every 100 agent steps, the trajectory of experience from each player's point of view  (observations, actions, rewards) is sent to the \emph{learner} responsible for the policy carried out by that player.
The learner corresponding to an agent composes batches of the 32 trajectories most recently received from arenas, and computes a weight update to the agent's neural network parameters based on \eqnref{eqn:hiervar} using V-Trace off-policy correction~\cite{espeholt2018impala} to account for off-policy drift.

\section{Performance Evaluation}
An important dimension of assessing the success of training agents to play CTF is to evaluate their skill in terms of the agent team's win probability. As opposed to single-agent tasks, assessing skill in multi-agent systems depends on the teammates and opponents used during evaluation. We quantified agent skill by playing evaluation games with players from the set of all agents to be assessed. Evaluation games were composed using ad-hoc matchmaking in the sense that all $N$ players of the game, from both teams, were drawn at random from the set of agents being evaluated. This allowed us to measure skill against any set of opponent agents and robustness to any set of teammate agents. We estimate skill using the Elo rating system~\cite{ELO} extended to teams
(see Section~\ref{sec:supp_elo} for exact details of Elo calculation).

We performed evaluation matches with snapshots of the FTW agent and ablation study agents through training time, and also included \emph{built-in bots} and \emph{human participants} as reference agents for evaluation purposes only. Differences between these types of players is summarised in Figure~\ref{fig:humanagentdiff}.

The various ablated agents in experiments are (i) UNREAL~\cite{JaderbergUnreal} trained with self-play using game winning reward -- this represents the state-of-the-art naive baseline -- (ii) Self-play with reward shaping (RS) which instead uses the Quake default points scheme as reward, (iii) PBT with RS, which replaces self-play with population based training, and (iv) FTW without temporal hierarchy which is the full FTW agent but omitting the temporal hierarchy (Section~\ref{sec:ablation} for full details).

The built-in bots were scripted AI bots developed for Quake III Arena. Their policy has access to the entire game engine, game state, and map layout, but have no learning component~\cite{waveren2001quakebots}. These bots were configured for various skill levels, from Bot 1 (very low skill level) to Bot 5 (very high skill level, increased shields), as described fully in Section~\ref{sec:botdetails}.

The human participants consisted of 40 people with first-person video game playing experience. We collected results of evaluation games involving humans by playing five tournaments of eight human players. Players were given instructions on the game environment and rules, and performed two games against Bot 3 built-in bots. Human players then played seven games in ad-hoc teams, being randomly matched with other humans, FTW agents, and FTW without a temporal hierarchy agents as teammates and opponents.
Players were not told with which agent types they were playing and were not allowed to communicate with each other.
Agents were executed in real-time on the CPUs of the same workstations used by human players (desktops with a commodity GPU) without adversely affecting the frame-rate of the game.

\begin{figure}[t]
\centering
\includegraphics[height=1cm]{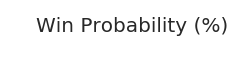}\\
\includegraphics[height=9cm]{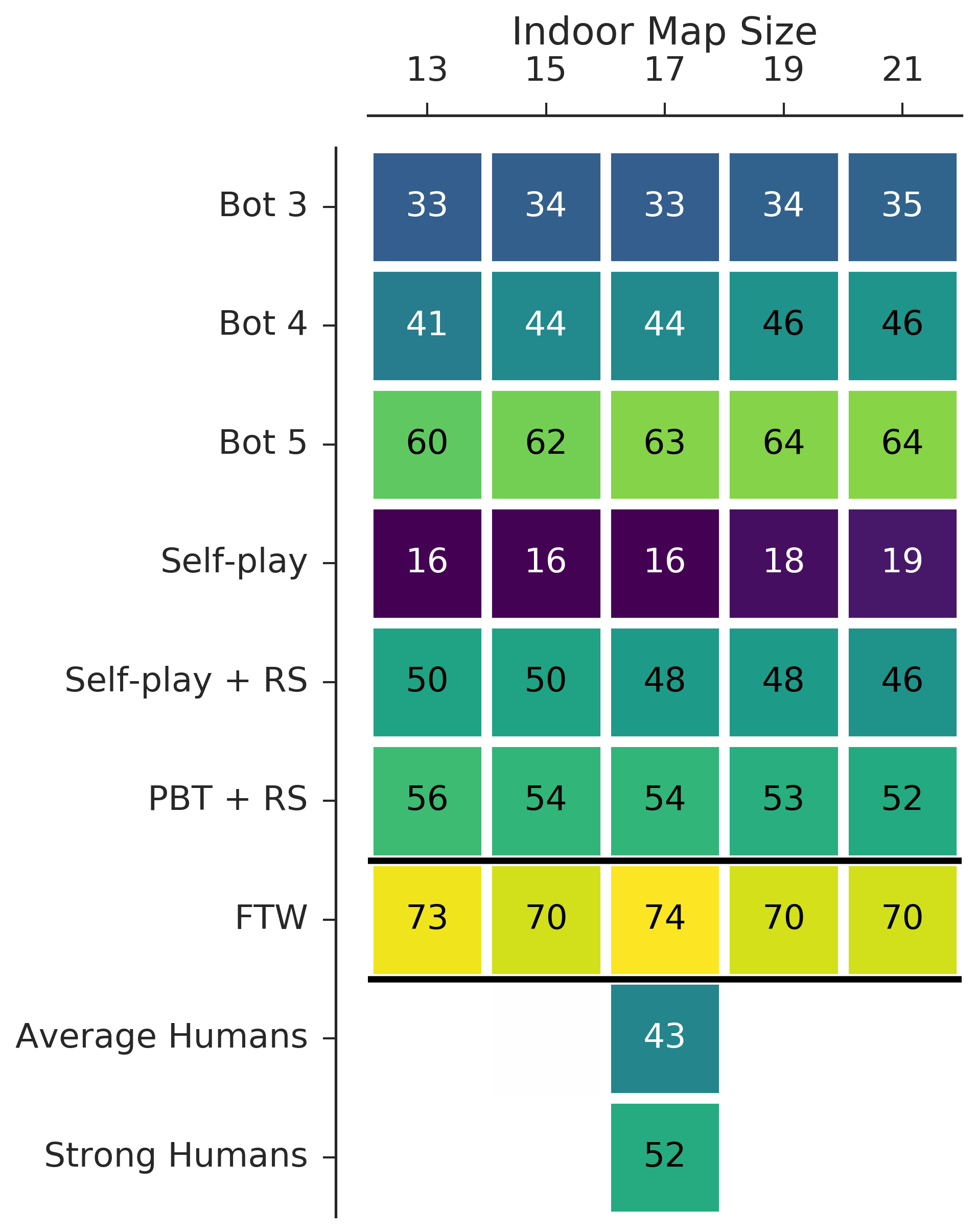}
\includegraphics[height=9cm]{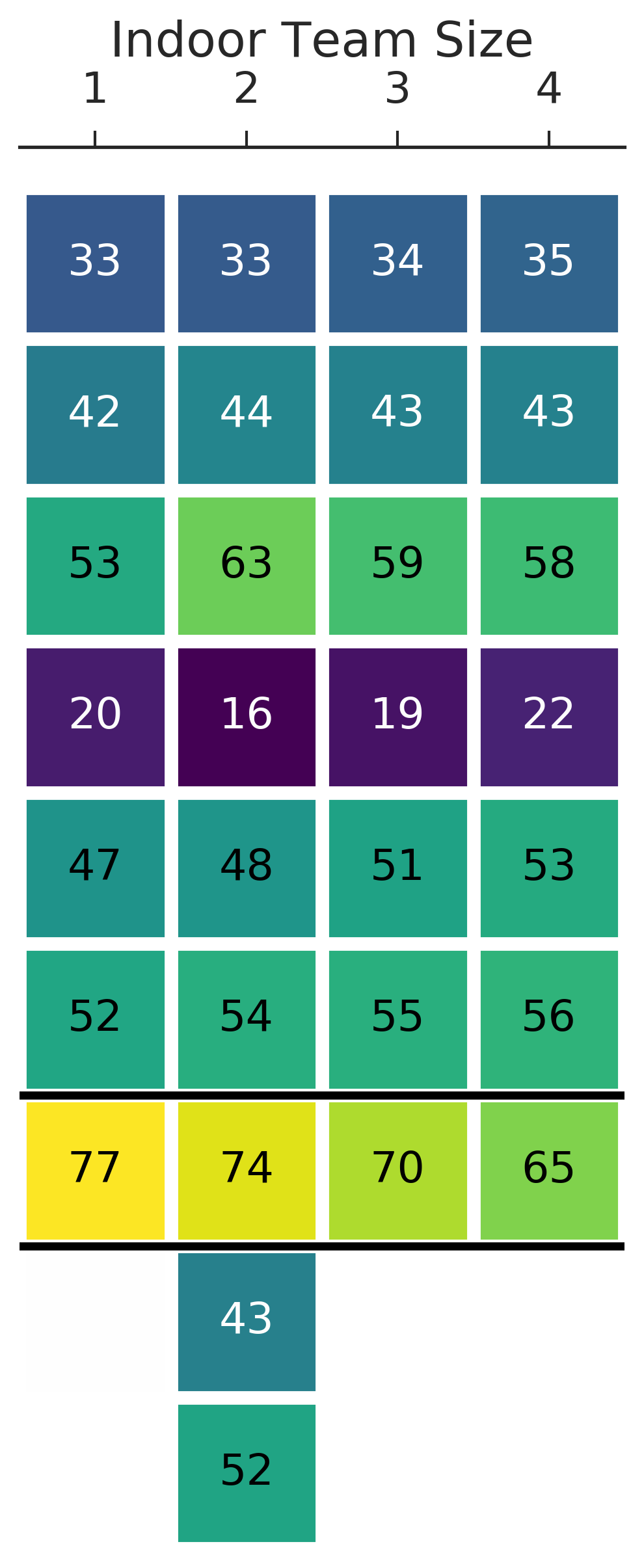}
\includegraphics[height=9cm]{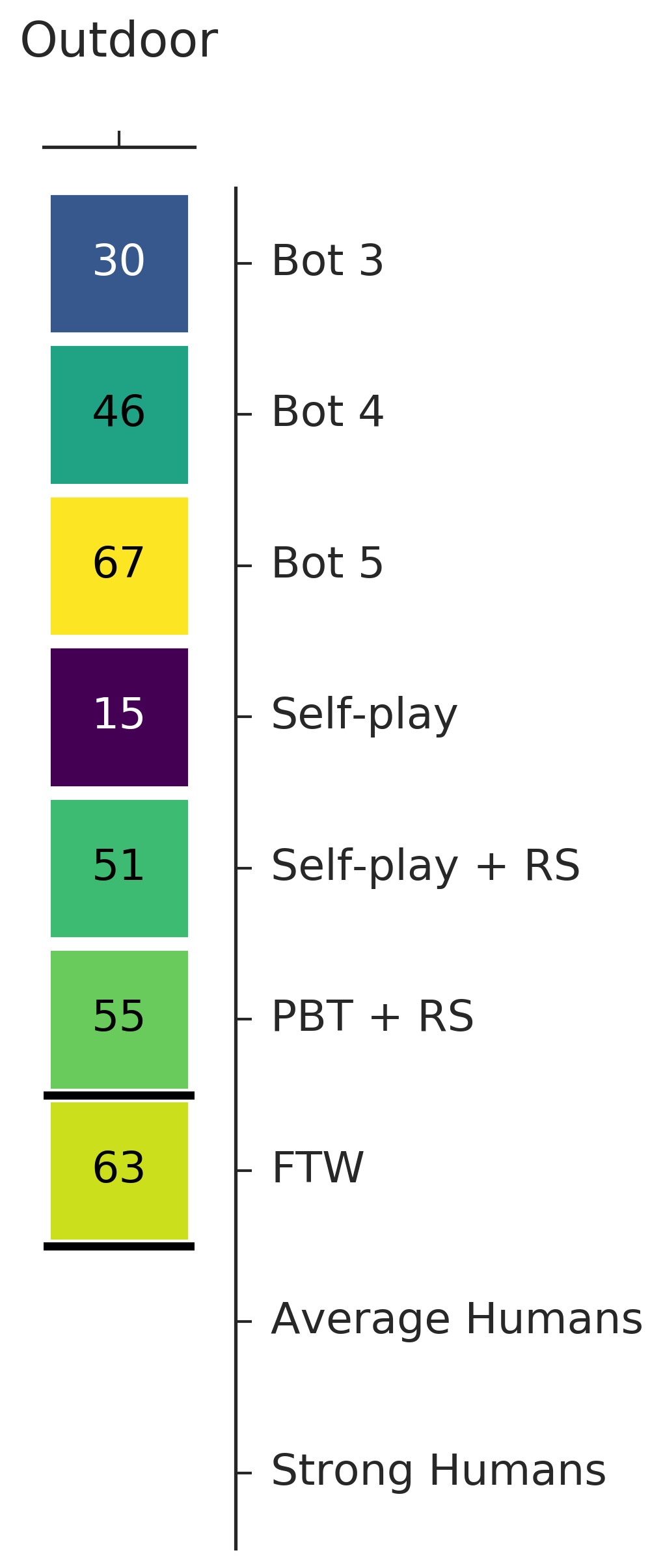}\\ \vspace{1cm}
\includegraphics[height=4.125cm]{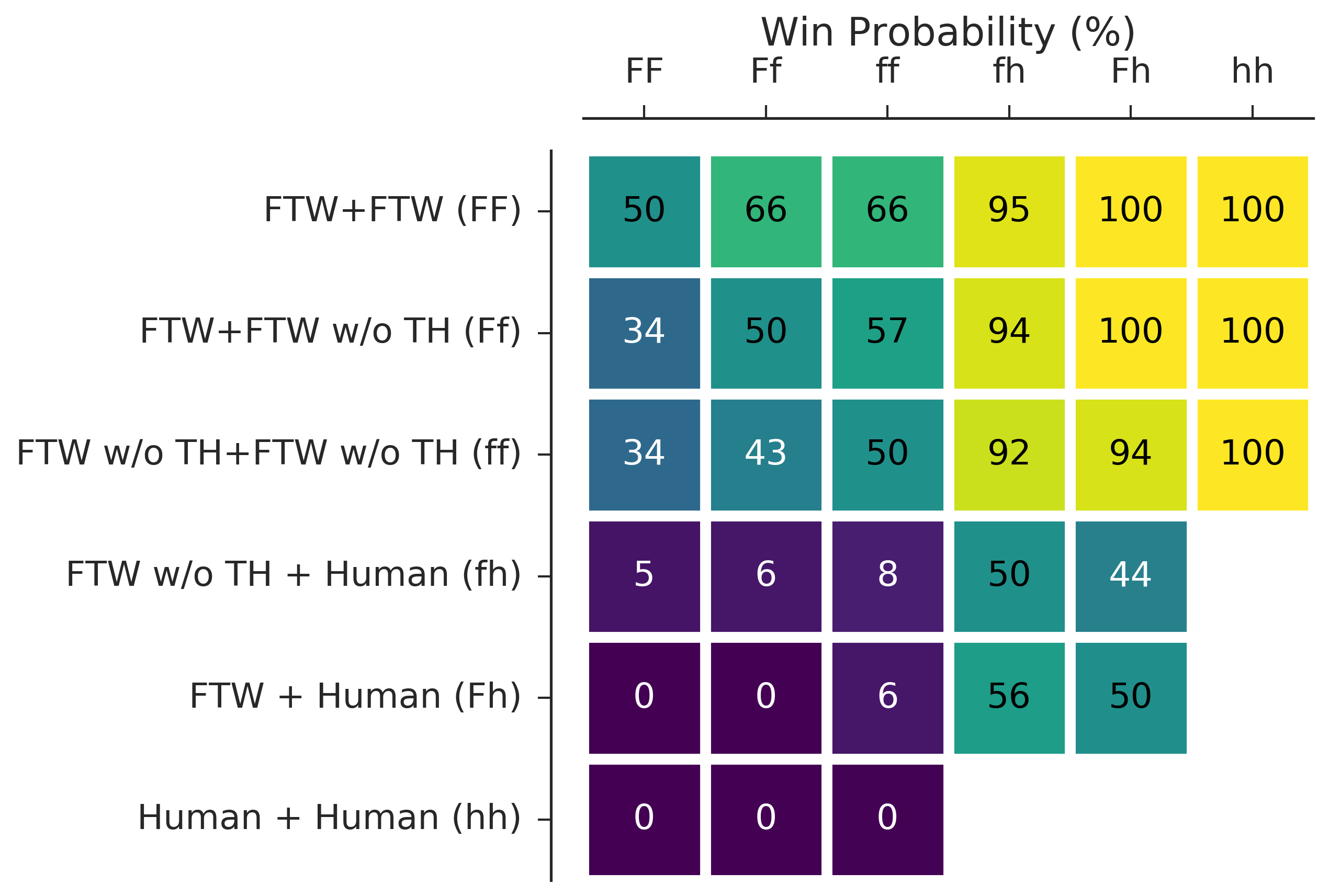}
\includegraphics[height=4.125cm]{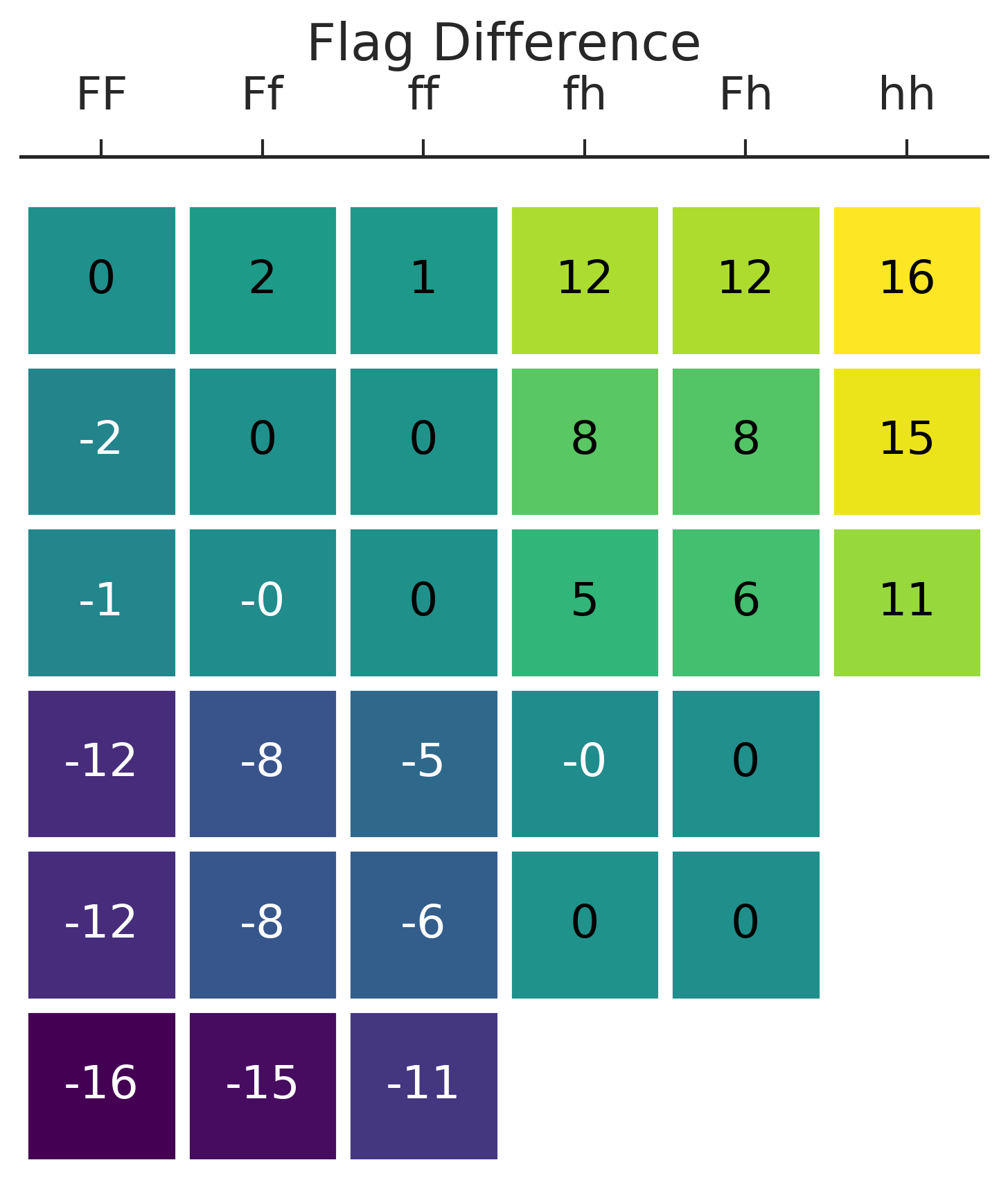}
\includegraphics[height=4.125cm]{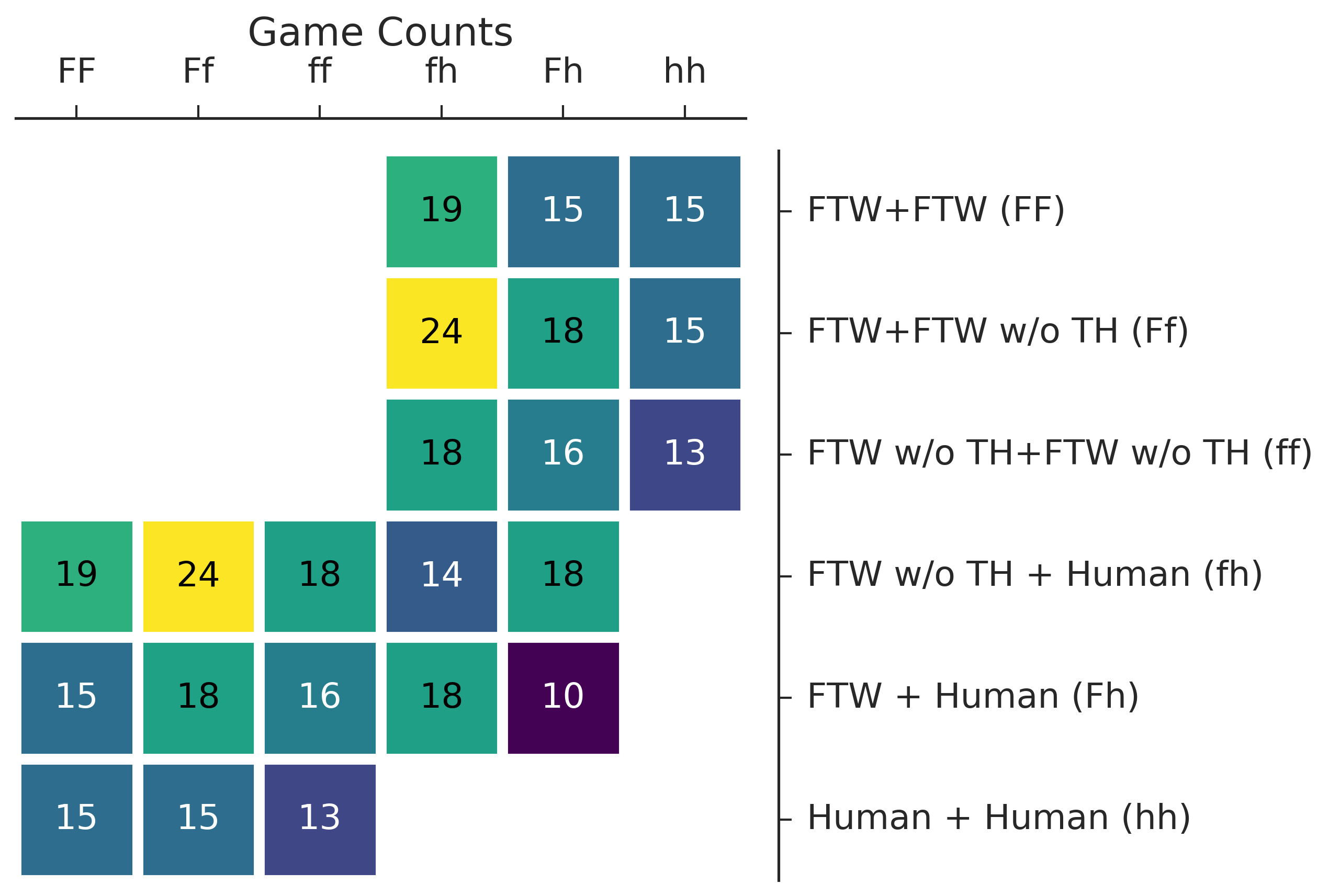}
\caption{\small
{\bf Top:} Shown are win probabilities of different agents, including bots and humans, in evaluation tournaments, when playing on procedurally generated maps of various sizes (13--21), team sizes (1--4) and styles (indoor/outdoor). On indoor maps, agents were trained with team size two on a mixture of $13\times 13$ and $17 \times 17$ maps, so performance in scenarios with different map and team sizes measures their ability to successfully generalise. Teams were composed by sampling agents from the set in the figure with replacement. {\bf Bottom: } Shown are win probabilities, differences in number of flags captured, and number of games played for the human evaluation tournament, in which human subjects played with agents as teammates and/or opponents on indoor procedurally generated $17 \times 17$ maps.}
\label{tab:leaderboard_proc}
\label{tab:leaderboard_natlab}
\label{tab:leaderboard_proc_nsvsn}
\label{fig:ext_human_win_prob}
\end{figure}

Figure \ref{fig:ext_human_win_prob} shows the outcome of the tournaments involving humans. To obtain statistically valid Elo estimates from the small number of games played among individuals with high skill variance, we pooled the humans into two groups, top 20\% (strong) and remaining 80\% (average), according to their individual performances.

\begin{figure}
\centering
\begin{tabular}[t]{cc}
\begin{tabular}[b]{lr}
\multicolumn{2}{c}{\bf Two-player Fetch}\\
\hline
Agent           &  Flags \\
\hline
Bot             &  14 \\
Self-play + RS &    9 \\
PBT + RS       &   14 \\
FTW w/o TH      &  23 \\
FTW             &  37 \\
\hline
Fetch-trained FTW  w/o TH & 30\\
Fetch-trained FTW & 44\\
\hline
\end{tabular}&
\begin{tabular}[b]{l}
\includegraphics[width=0.195\textwidth]{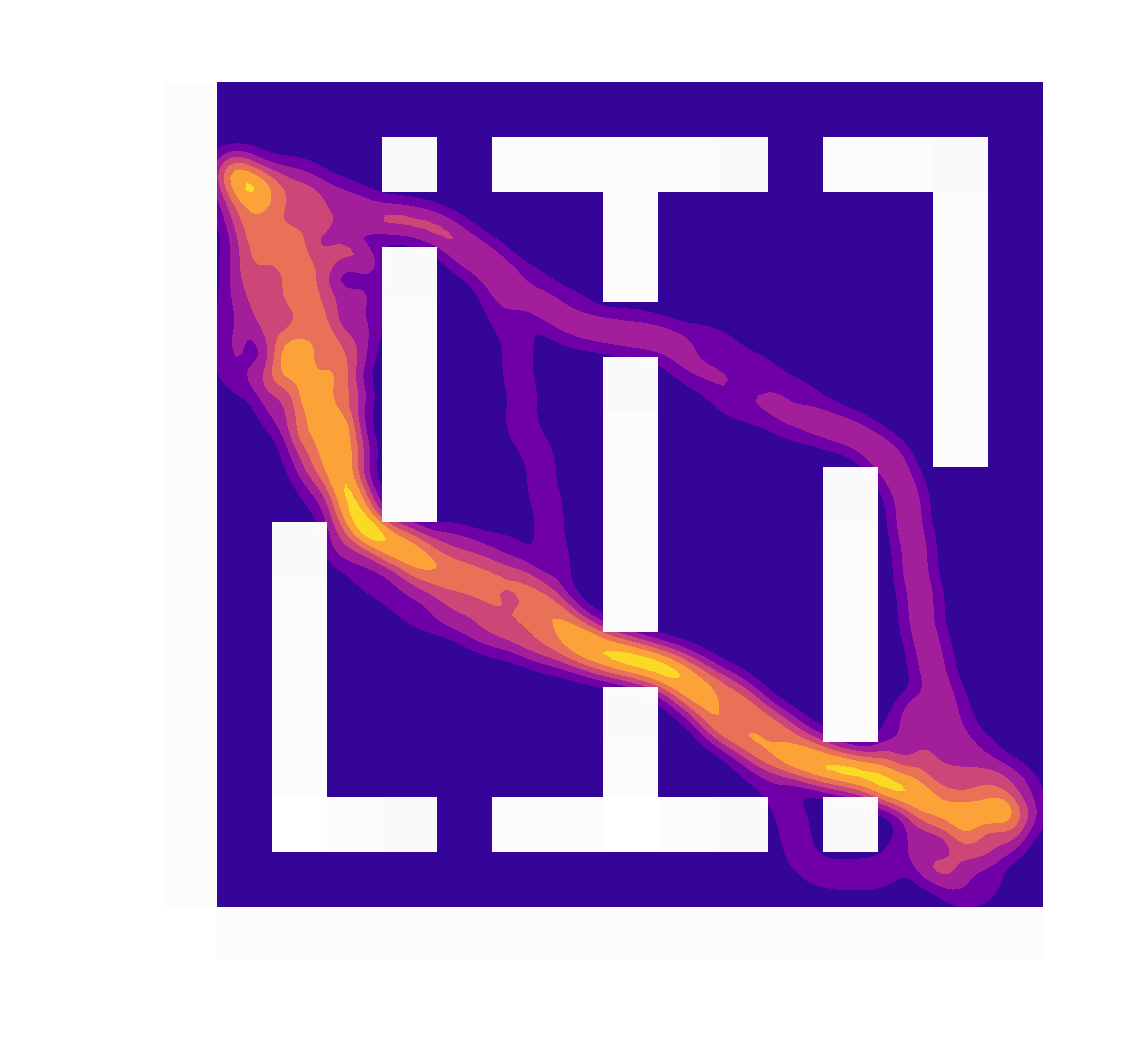}
\includegraphics[width=0.195\textwidth]{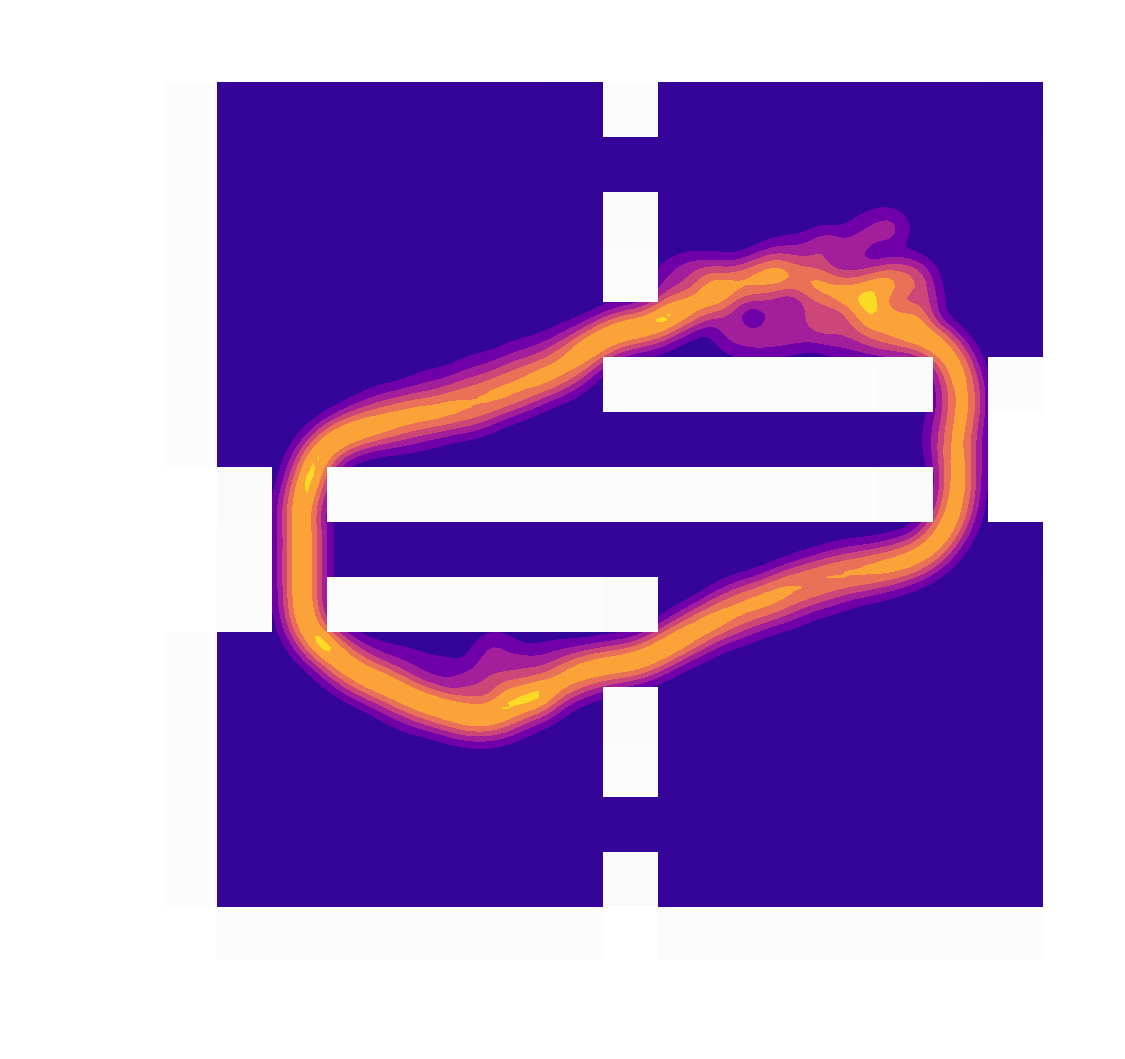}
\includegraphics[width=0.195\textwidth]{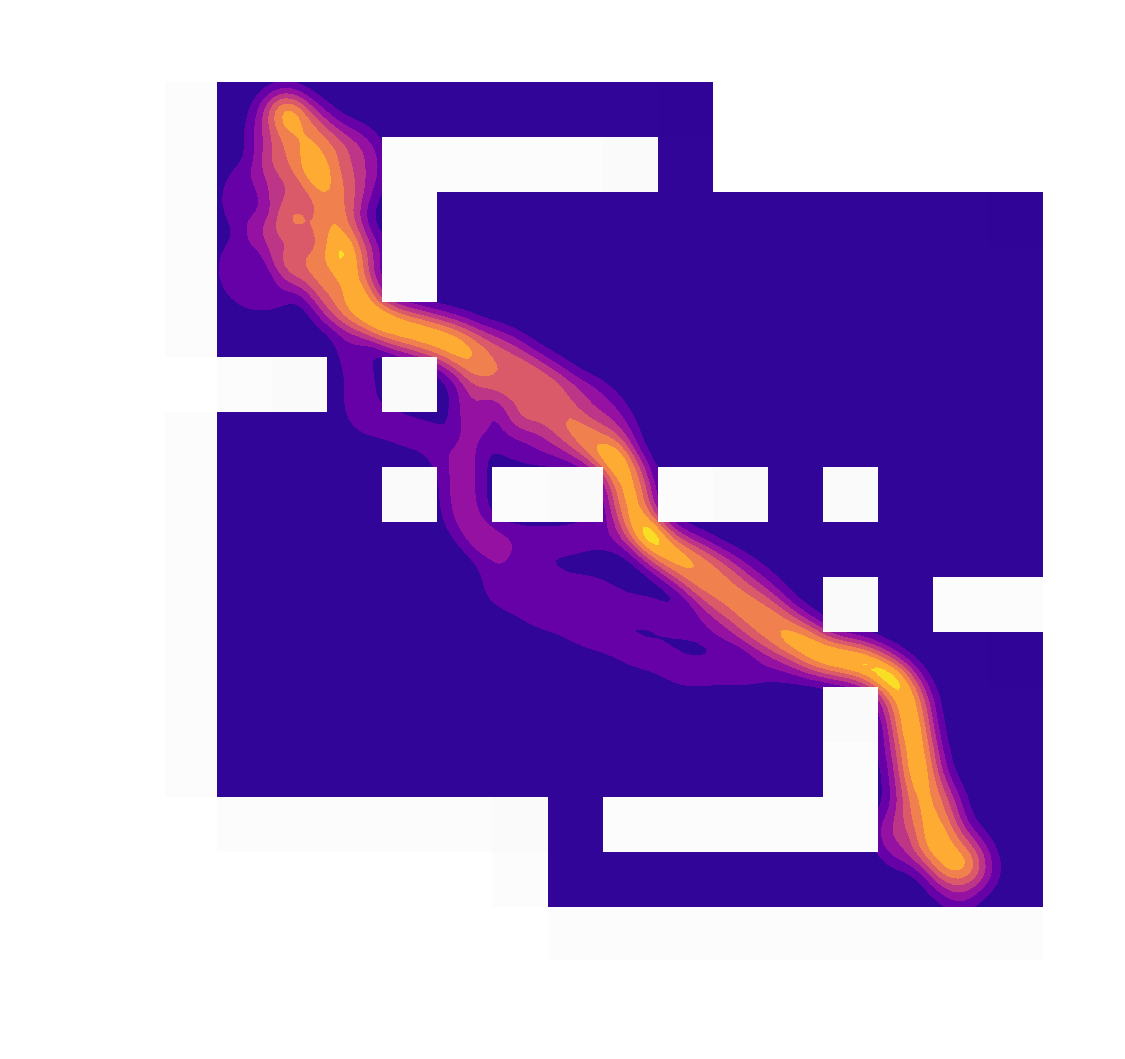}\\
\includegraphics[width=0.195\textwidth]{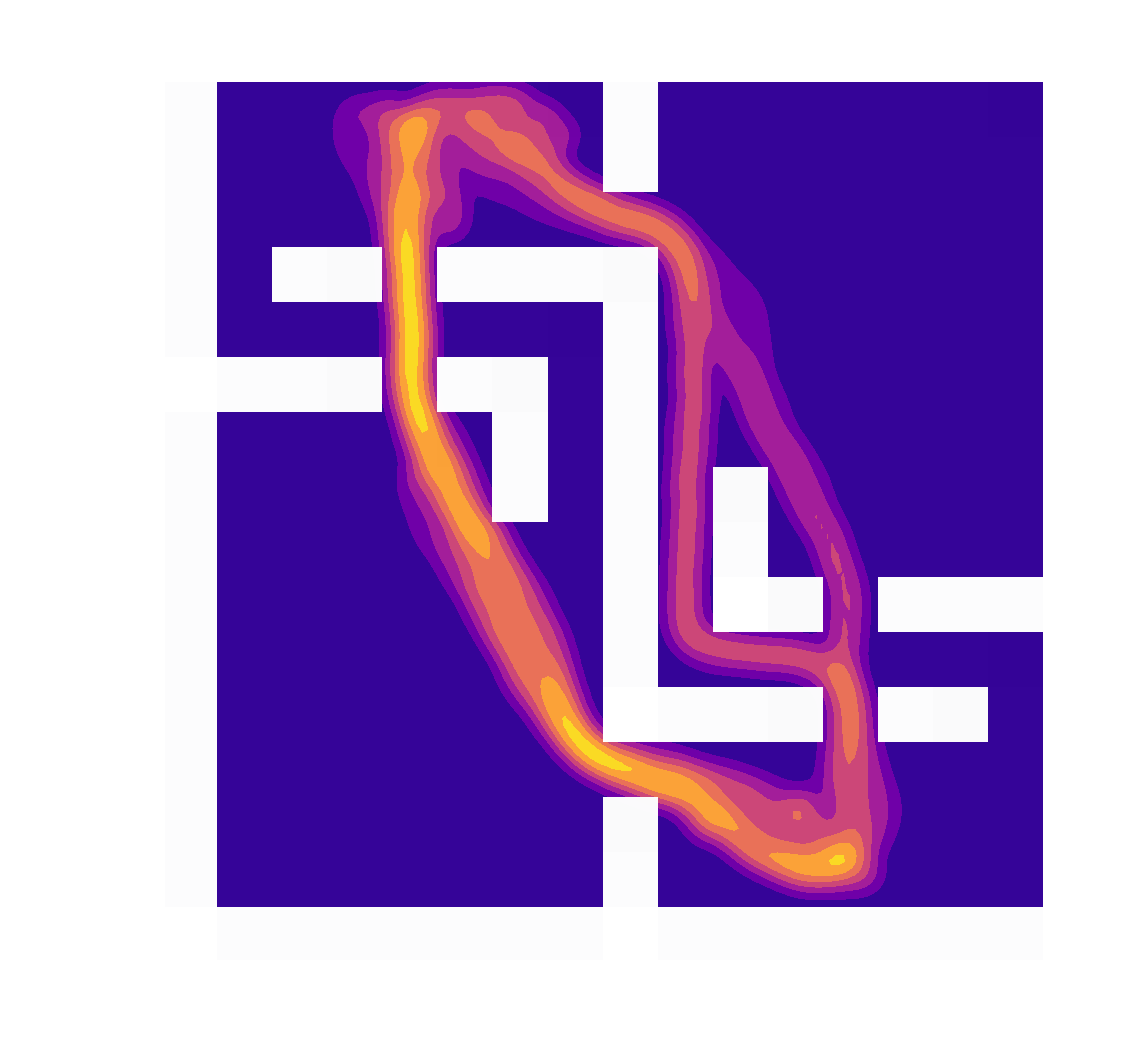}
\includegraphics[width=0.195\textwidth]{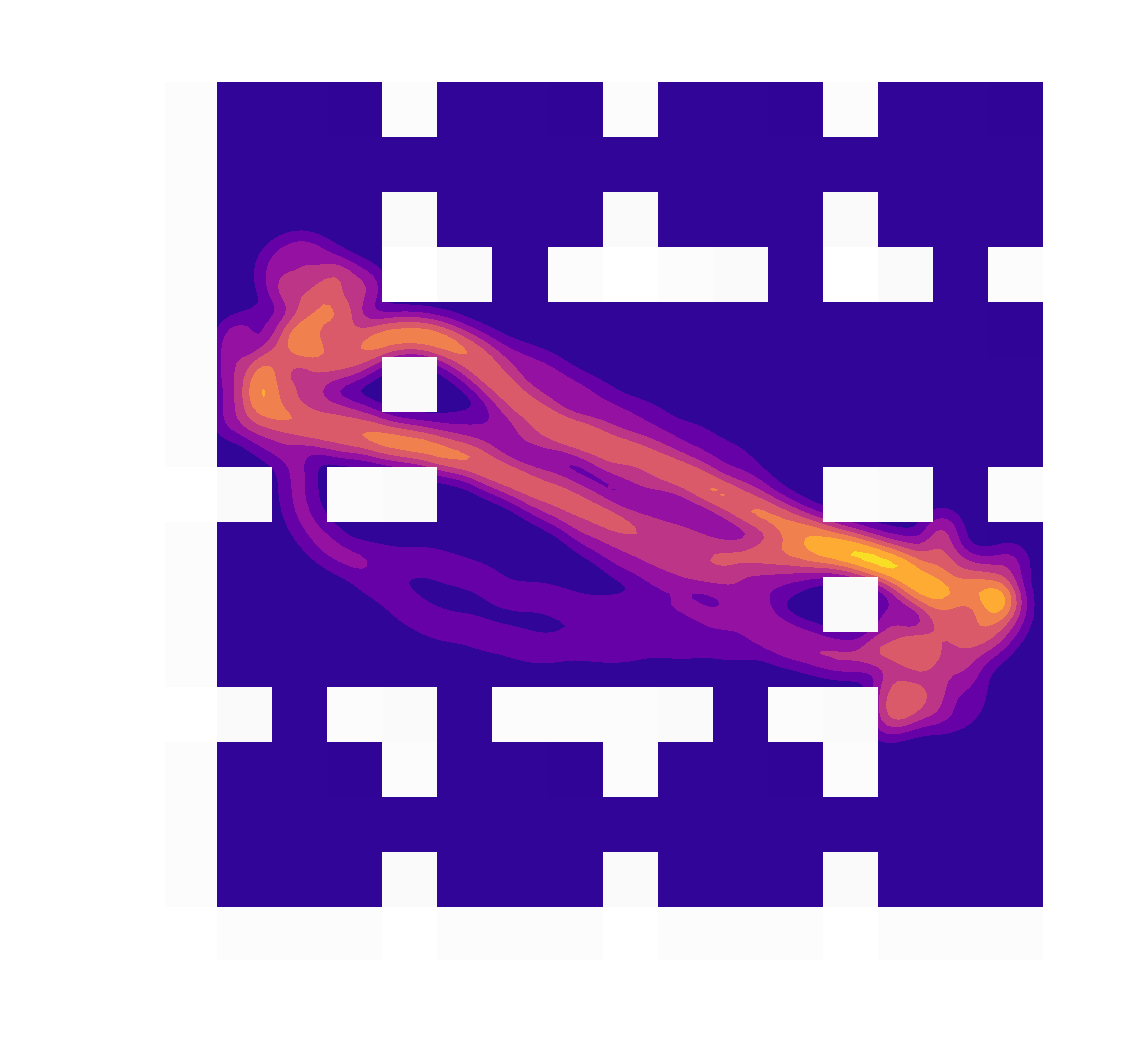}
\includegraphics[width=0.195\textwidth]{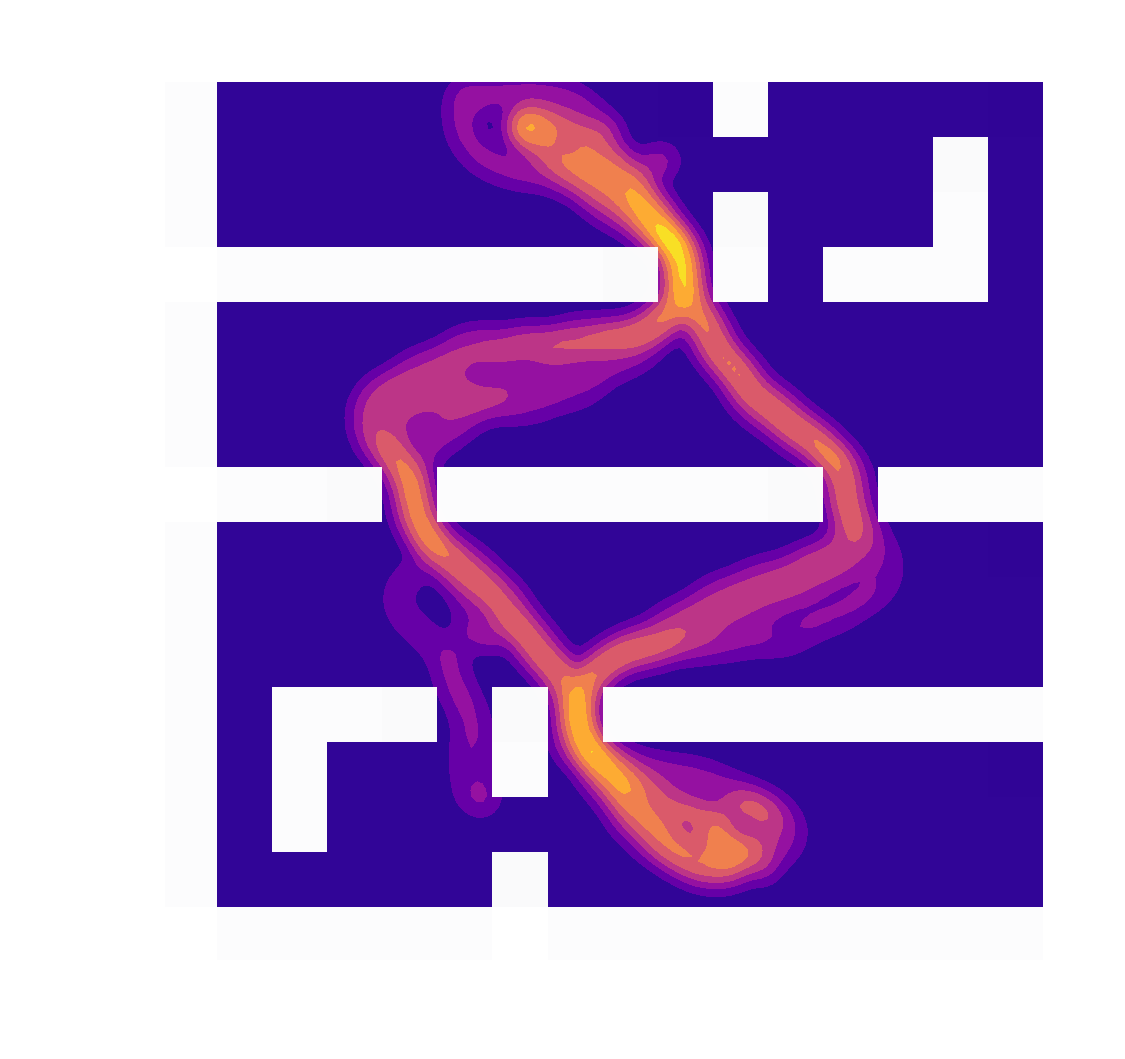}
\end{tabular}
\end{tabular}
\caption{\small
{\bf Left:} Average number of flags scored per match for different CTF-trained agents playing two-player \emph{fetch} (CTF without opponents) on indoor procedurally generated maps of size 17. This test provides a measure of agents' ability to cooperate while navigating in previously unseen maps.
Ten thousand matches were played, with teams consisting of two copies of the same agent, which had not been trained on this variant of the CTF task.
All bot levels performed very similarly on this task, so we report a single number for all bot levels. In addition we show results when agents are trained solely on the fetch task (+1 reward for picking up and capturing a flag only).
{\bf Right:} Heatmaps of the visitation of the FTW agent during the second half of several episodes while playing fetch..}
\label{fig:fetch}
\label{tab:leaderboard_fetch_2vs2}
\end{figure}

We also performed another study with human players to find out if human ingenuity, adaptivity and teamwork would help humans find exploitative strategies against trained agents. We asked two professional games testers to play as a team against a team of two FTW agents on a fixed, particularly complex map, which had been held out of training. After six hours of practice and experimentation, the human games testers were able to consistently win against the FTW team on this single map by employing a high-level strategy. This winning strategy involved careful study of the preferred routes of the agents on this map in exploratory games, drawing explicit maps, and then precise communication between humans to coordinate successful flag captures by avoiding the agents' preferred routes.
In a second test, the maps were changed to be procedurally generated for each episode as during training. Under these conditions, the human game testers were not able to find a consistently winning strategy, resulting in a human win-rate of only 25\% (draw rate of 6.3\%).

\subsection{Human-Agent Differences}\label{sec:humandiff}

\begin{figure}
    \centering
    \includegraphics[width=\textwidth]{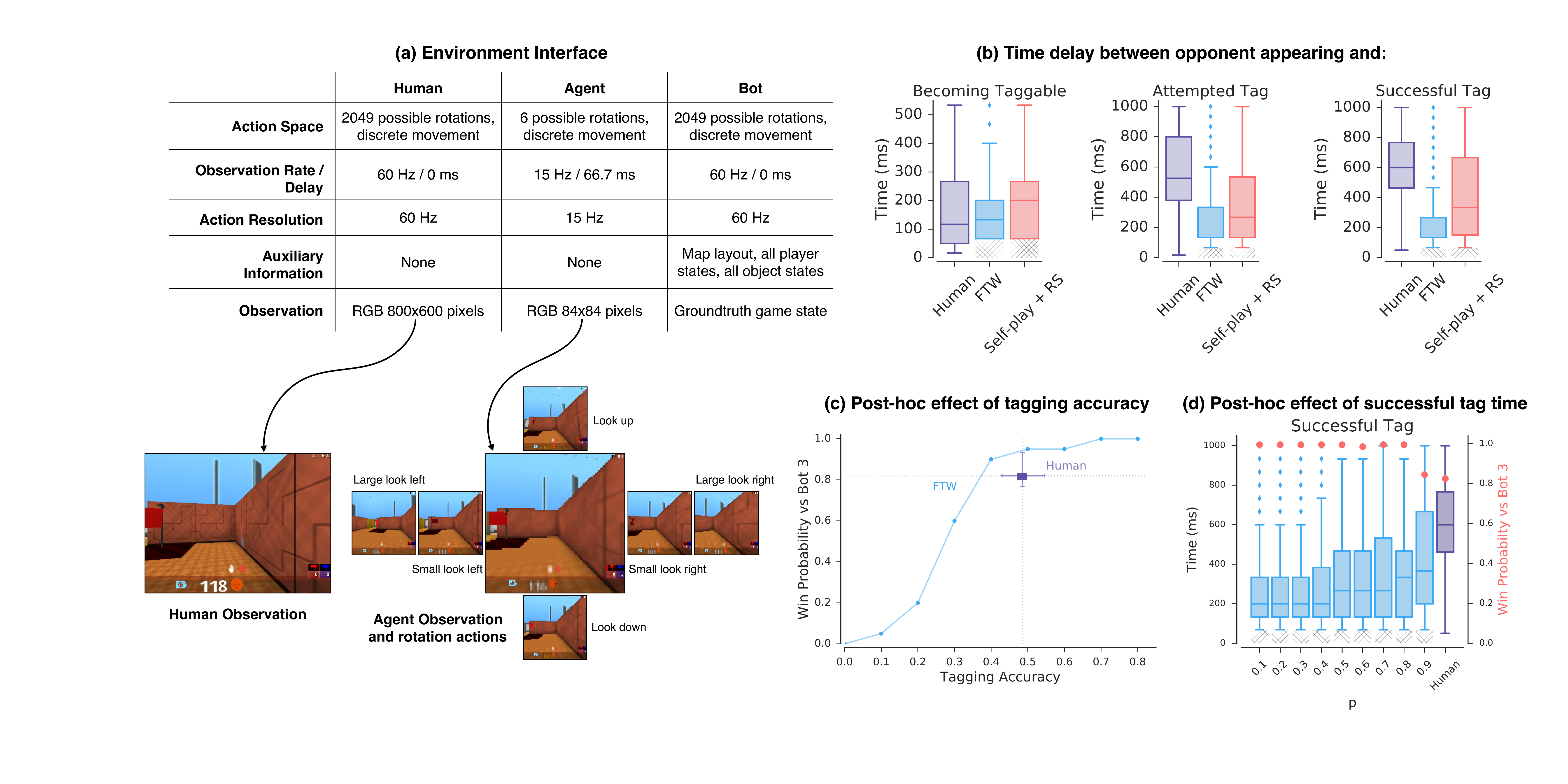}
    \caption{
    \small {\bf (a)} The differences between the environment interface offered to humans, agents, and bots.
    {\bf (b)} Humans and agents are in addition bound by other sensorimotor limitations. To illustrate we measure humans' and agents' response times, when playing against a Bot 3 team on indoor procedural maps. Time delays all measured from first appearance of an opponent in an observation. Left: delay until the opponent becoming taggable (\ie~lies within a 10 degree visual cone). Middle: delay until an attempted tag (\ie~opponent lies within a 10 degree visual cone and tag action emitted). Right: delay until a successful tag. We ignore situations where opponents are further than 1.5 map units away. The shaded region represents values which are impossible to obtain due to environment constraints.
    {\bf (c)} Effect of tagging accuracy on win probability against a Bot 3 team on indoor procedural maps. Accuracy is the number of successful tags divided by valid tag attempts. Agents have a trained accuracy of 80\%, much higher than the 48\% of humans. In order to measure the effect of decreased accuracy on the FTW agent, additional evaluation matches were performed where a proportion of tag events were artificially discarded. As the agent's accuracy increases from below human (40\%) to 80\% the win probability increases from 90\% to 100\% which represents a significant change in performance.
    {\bf (d)} Effect of successful tag time on win probability against a Bot 3 team on indoor procedural maps. In contrast to (c), the tag \emph{actions} were artificially discarded $p$\% of the time -- different values of $p$ result in the spectrum of response times reported. Values of $p$ greater than 0.9 did not reduce response time, showing the limitations of $p$ as a proxy.
    Note that in both (c) and (d), the agents were not retrained with these $p$ values and so obtained values are only a lower-bound of the potential performance of agents -- this relies on the agents generalising outside of the physical environment they were trained in. }
    \label{fig:humanagentdiff}
\end{figure}

It is important to recognise the intrinsic differences between agents and humans when evaluating results. It is very difficult to obtain an even playing ground between humans and agents, and it is likely that this will continue to be the case for all human/machine comparisons in the domain of action video games. While we attempted to ensure that the interaction of agents and humans within their shared environment was as fair as possible, engineering limitations mean that differences still exist. Figure~\ref{fig:humanagentdiff} (a) outlines these, which include the fact that the environment serves humans a richer interface than agents: observations with higher visual resolution and lower temporal latency, and a control space of higher fidelity and temporal resolution.

However, in spite of these environmental constraints, agents have a set of advantages over humans in terms of their ultimate sensorimotor precision and perception. Humans cannot take full advantage of what the environment offers: they have a visual-response feedback loop far slower than the 60Hz observation rate~\cite{castel2005effects}; and although a high fidelity action space is available, humans' cognitive and motor skills limit their effective control in video games~\cite{berard2011limits}.

One way that this manifests in CTF games is through reaction times to salient events. While we cannot measure reaction time directly within a full CTF game, we measure possible proxies for reaction time by considering how long it takes for an agent to respond to a newly-appeared opponent (Figure~\ref{fig:humanagentdiff} (b)). After an opponent first appears within a player's (90 degree) field-of-view, it must be become ``taggable'', \ie~positioned within a 10 degree cone of the player's centre of vision. This occurs very quickly within both human and agent play, less than 200ms on average (though this does not necessarily reflect intentional reactions, and may also result from some combination of players' movement statistics and prior orientation towards opponent appearance points). However, the time between first seeing an opponent and attempting a tag (the opponent is taggable and the tag action is emitted) is much lower for FTW agents (258ms on average) compared to humans (559ms), and when a successful tag is considered this gap widens (233ms FTW, 627ms humans). Stronger agents also had lower response times in general than weaker agents, but there was no statistically significant difference in strong humans' response times compared to average humans.

The tagging accuracy of agents is also significantly higher than that of humans: 80\% for FTW agents compared to 48\% for humans. We measured the effect of tagging accuracy on the performance of FTW agents playing against a Bot 3 team by artificially impairing agents’ ability to fire, without retraining the agents (Figure~\ref{fig:humanagentdiff} (c)). Win probability decreased as the accuracy of the agent decreased, however at accuracies comparable to humans the FTW agents still had a greater win probability than humans (albeit with comparable mean flag capture differences). We also used this mechanism to attempt to measure the effect of successful tag time on win probability (Figure~\ref{fig:humanagentdiff} (d)), and found that an average response time of up to 375ms did not effect the win probability of the FTW agent -- only at 448ms did the win rate drop to 85\%.

\section{Analysis}
\subsection{Knowledge Representation}
We carried out an analysis of the FTW agent's internal representation to help us understand how it represents its environment, what aspects of game state are well represented, and how it uses its memory module and parses visual observations.

\begin{figure}[t]
    \centering
    \includegraphics[width=\textwidth]{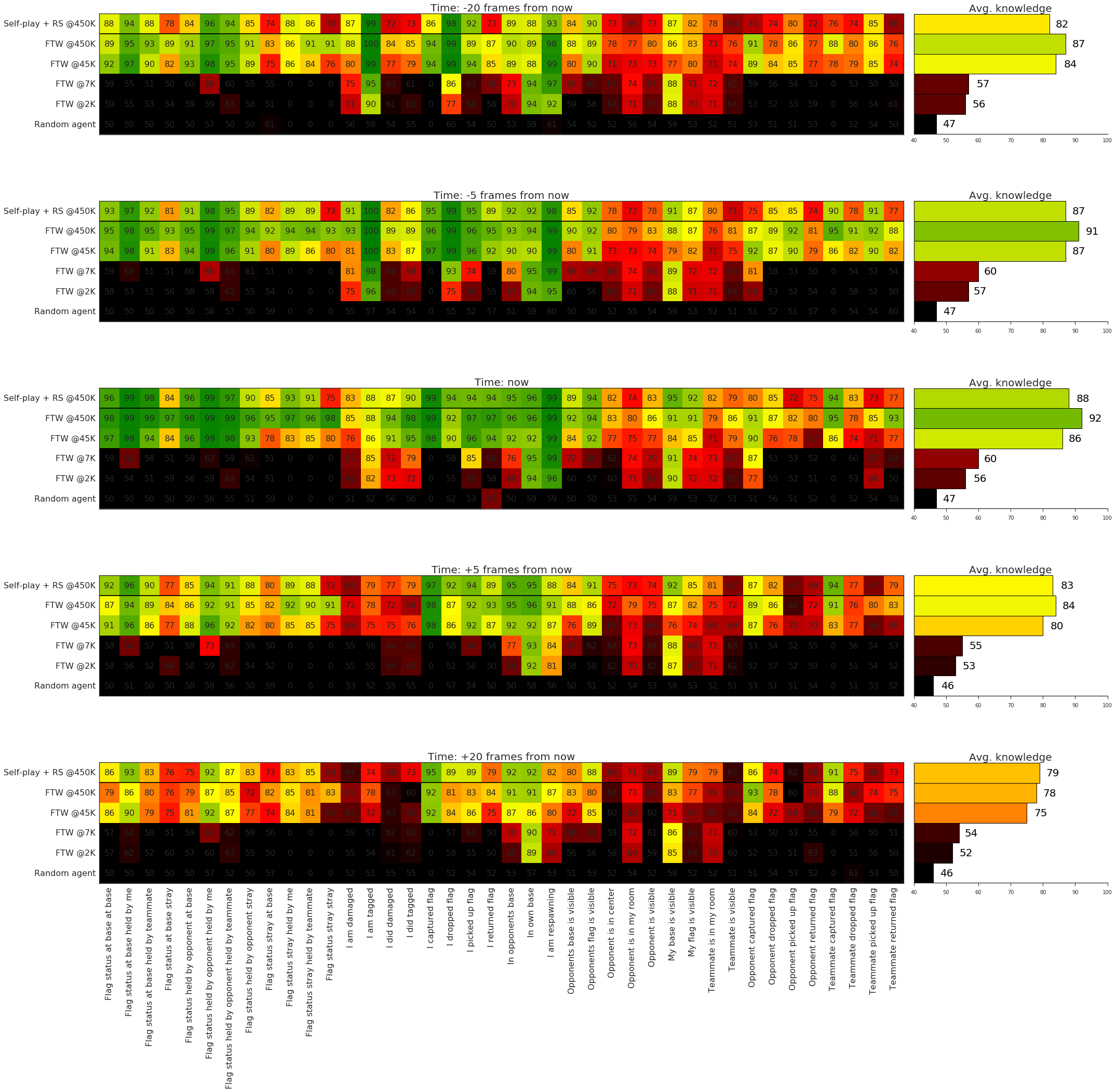}
    \caption{\small Shown is prediction accuracy in terms of percent AUCROC of linear probe classifiers on 40 different high-level game state features (columns) for different agents (rows), followed by their averages across features, for five different temporal offsets ranging from -20 to +20 frames (top to bottom). Results are shown for the baseline self-play agent with reward shaping as well as the FTW agent after different numbers of training games, and an untrained randomly initialised FTW agent.}
    \label{fig:ext_knowledge}
\end{figure}

We say that the agent had game-state related knowledge of a given piece of information if that information could be decoded with sufficient accuracy from the agent's recurrent hidden state $(\vec{h}_t^p, \vec{h}_t^q)$ using a linear probe classifier.
We defined a set of 40 binary features that took the form of questions (found in Figure~\ref{fig:ext_knowledge}) about the state of the game in the distant and recent past, present, and future, resulting in a total of 200 features. Probe classifiers were trained for each of the 200 features using balanced logistic regression on 4.5 million game situations, with results reported in terms of AUCROC evaluated with 3-fold episode-wise cross validation. This analysis was performed on the agent at multiple points in training to show what knowledge emerges at which point in training, with the results shown in Figure~\ref{fig:ext_knowledge}.

\begin{figure}[t]
    \centering
    \includegraphics[width=0.9\textwidth]{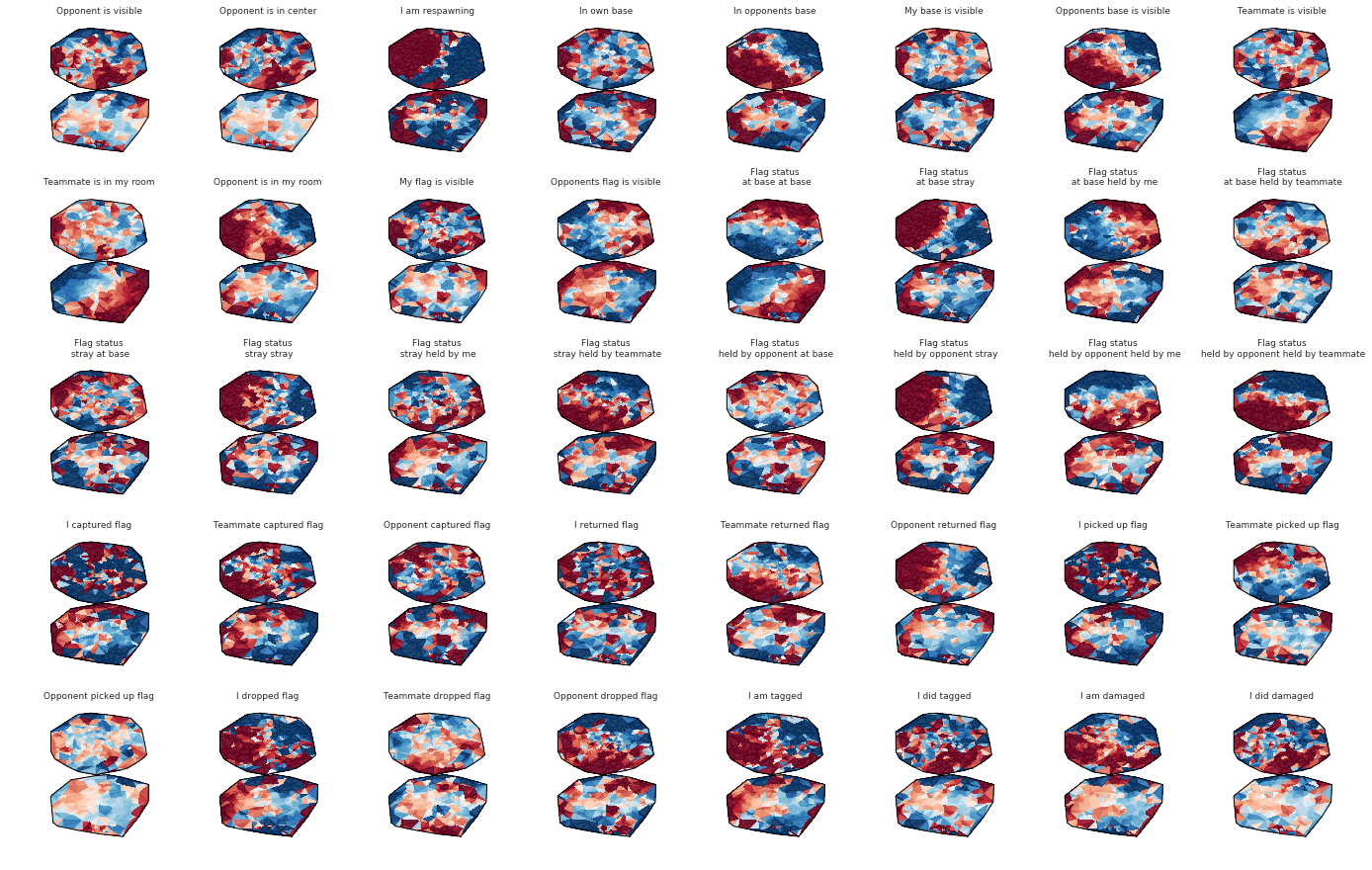}
    \includegraphics[width=0.6\textwidth]{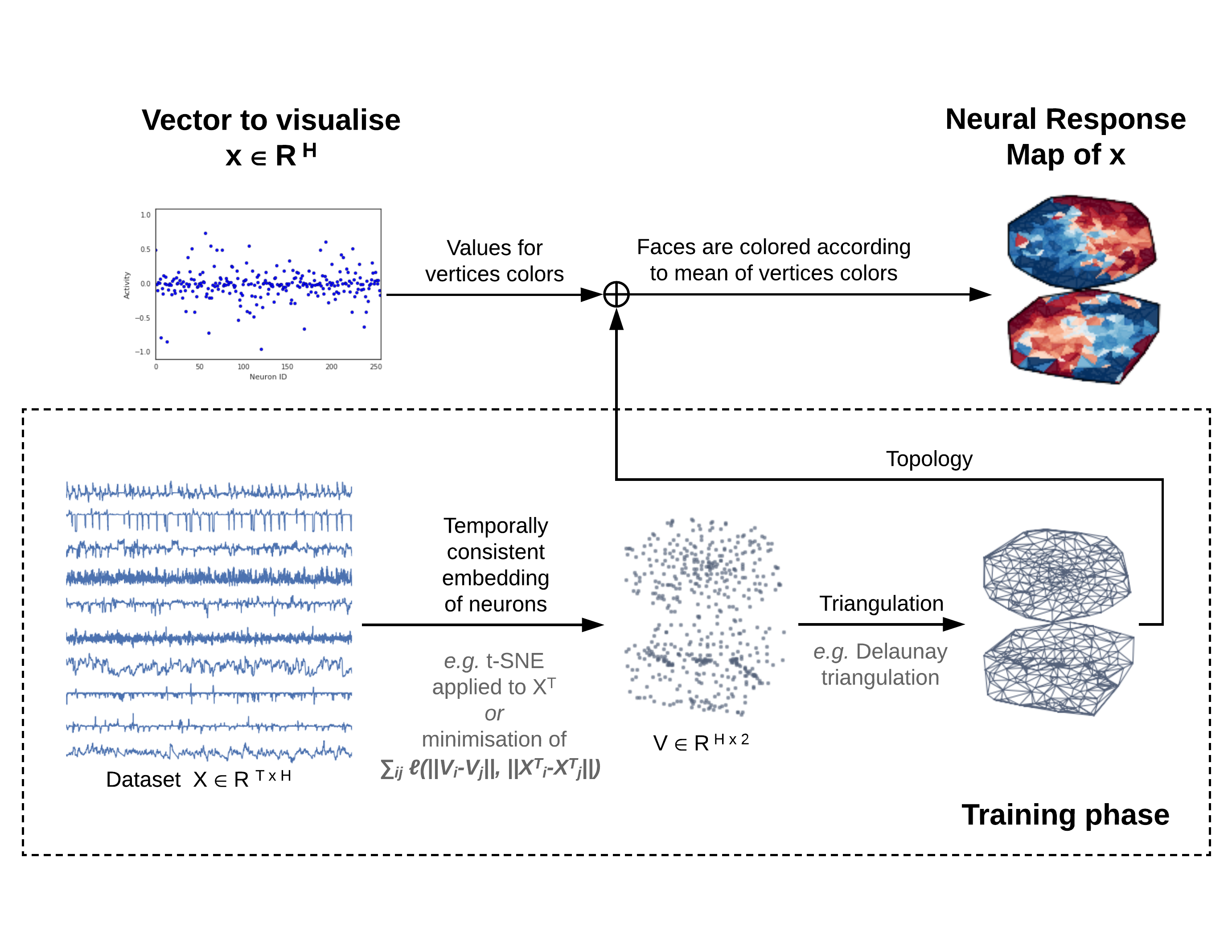}
    \caption{\small
    {\bf Top:} Shown are {\it neural response maps} for the FTW agent for game state features used in the knowledge study of Extended Data Figure~\ref{fig:ext_knowledge}. For each binary feature $y$ we plot the response vector $\mathbb{E}[(\vec{h}^p, \vec{h}^q) | y=1] - \mathbb{E}[(\vec{h}^p, \vec{h}^q) | y=0]$.
    {\bf Bottom:} Process for generating similarity based topological embedding of the elements of vector $\vec{x} \in \mathbb{R}^H$ given a dataset of other $X \in \mathbb{R}^{T \times H}$. Here we use two independent t-SNE embeddings, one for each of the agent's LSTM hidden state vectors at the two timescales.}
    \label{fig:ext_neural_response}
    \label{fig:ext_neural_responses}
\end{figure}

Further insights about the geometry of the representation space were gleaned by performing a t-SNE dimensionality reduction~\cite{maaten2008visualizing} on the recurrent hidden state of the FTW agent. We found strong evidence of cluster structure in the agent's representation reflecting conjunction of known CTF game state elements: flag possession, location of the agent, and the agent's respawn state. Furthermore, we introduce \emph{neural response maps} which clearly highlight the differences in co-activation of individual neurons of the agent in these different game states (Figure~\ref{fig:ext_neural_responses}). In fact, certain aspects of the game, such as whether an agent's flag is held by an opponent or not, or whether the agent's teammate holds the opponents flag or not, are represented by the response of single neurons.

\begin{figure}[t]
    \centering
    \includegraphics[width=0.92\textwidth]{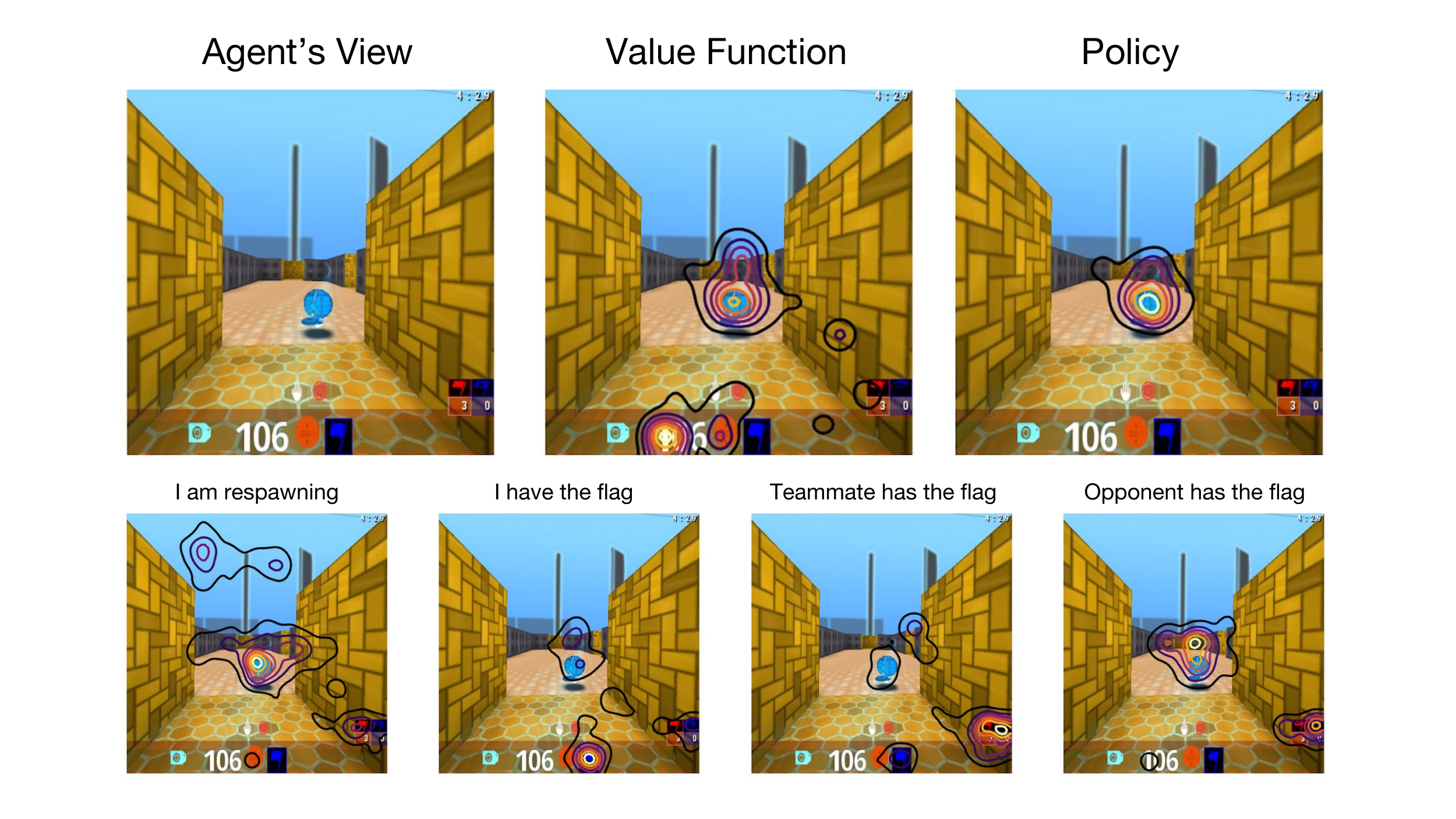}\\
    \includegraphics[width=0.8\textwidth]{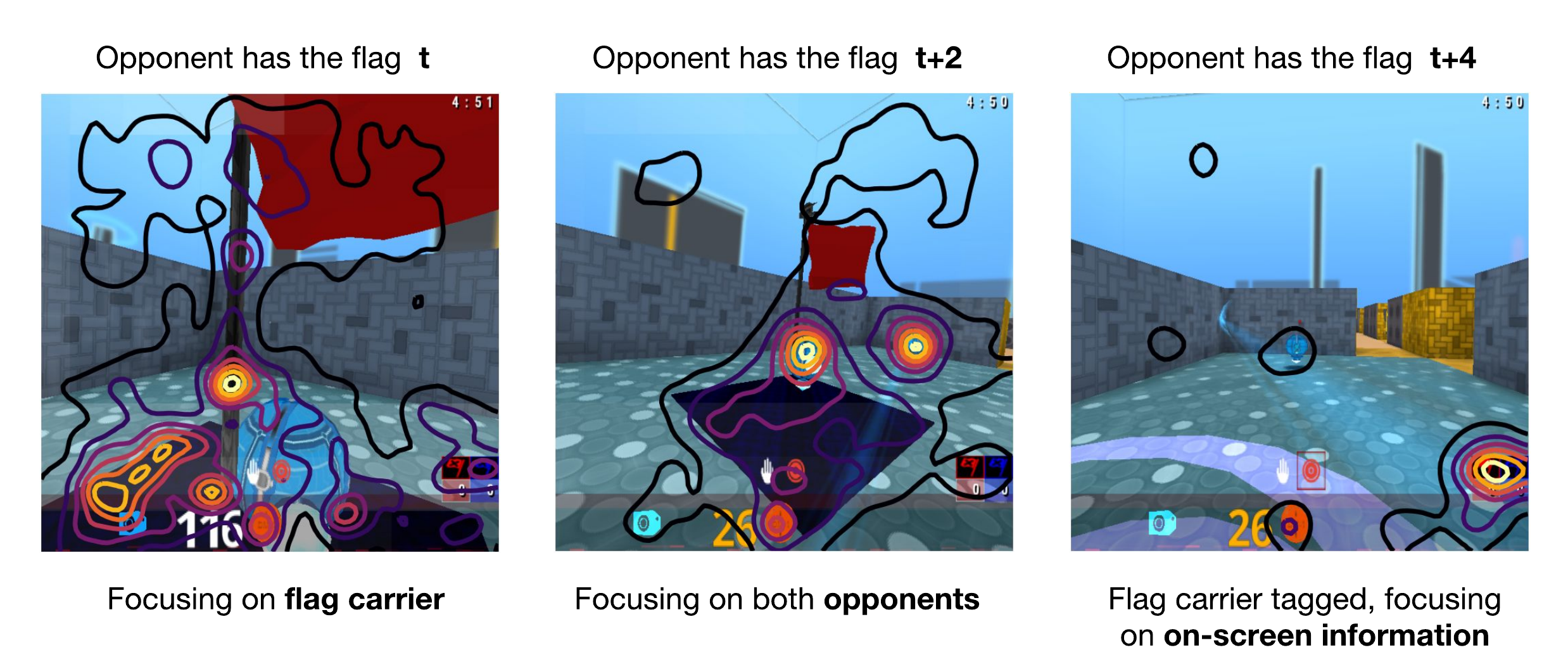}
    \caption{\small
    {\bf Top two rows:} Selected saliency analysis of FTW agent. Contours show sensitivity $\left \| \tfrac{\partial f_t}{\partial \vec{x}_{t,ij}} \right \|_1$, where $f_t$ is instantiated as the agent's value function at time $t$, its policy, or one of four highly selective neurons in the agent's hidden state, and $\vec{x}_{t,ij}$ represents the pixel at position $ij$ at time $t$. Brighter colour means higher gradient norm and thus higher sensitivity to given pixels.
    {\bf Bottom:} Saliency analysis of a single neuron that encodes whether an opponent is holding a flag. Shown is a single situation from the perspective of the FTW agent, in which attention is on an opponent flag carrier at time $t$, on both opponents at time $t+2$, and switches to the on-screen information at time $t+4$ once the flag carrier has been tagged and the flag returned.}
    \label{fig:ext_saliency}
    \label{fig:ext_attention}
\end{figure}

Finally, we can decode the sensitivity of the agent's value function, policy, and internal single neuron responses to its visual observations of the environment through gradient-based saliency analysis~\cite{simonyan2013deep} (Figure~\ref{fig:ext_saliency}). Sensitivity analysis combined with knowledge classifiers seems to indicate that the agent performed a kind of task-based scene understanding, with the effect that its value function estimate was sensitive to seeing the flag, other agents, and elements of the on-screen information. The exact scene objects which an agent's value function was sensitive to were often found to be context dependent (Figure~\ref{fig:ext_attention} bottom).

\begin{figure}[t]
    \centering
    \includegraphics[width=\textwidth]{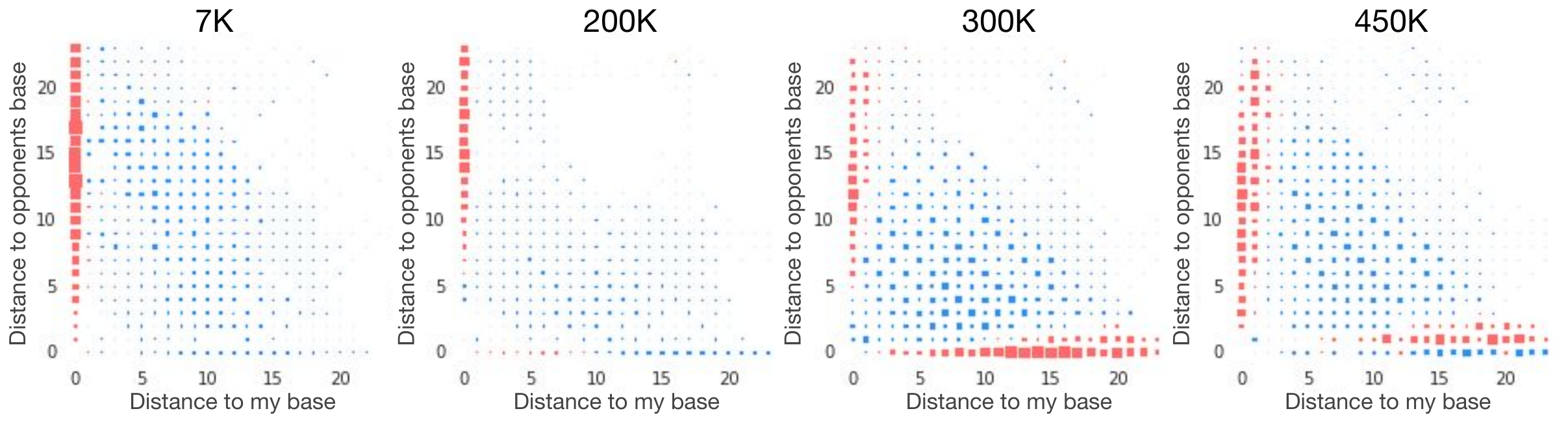}
    \includegraphics[width=\textwidth]{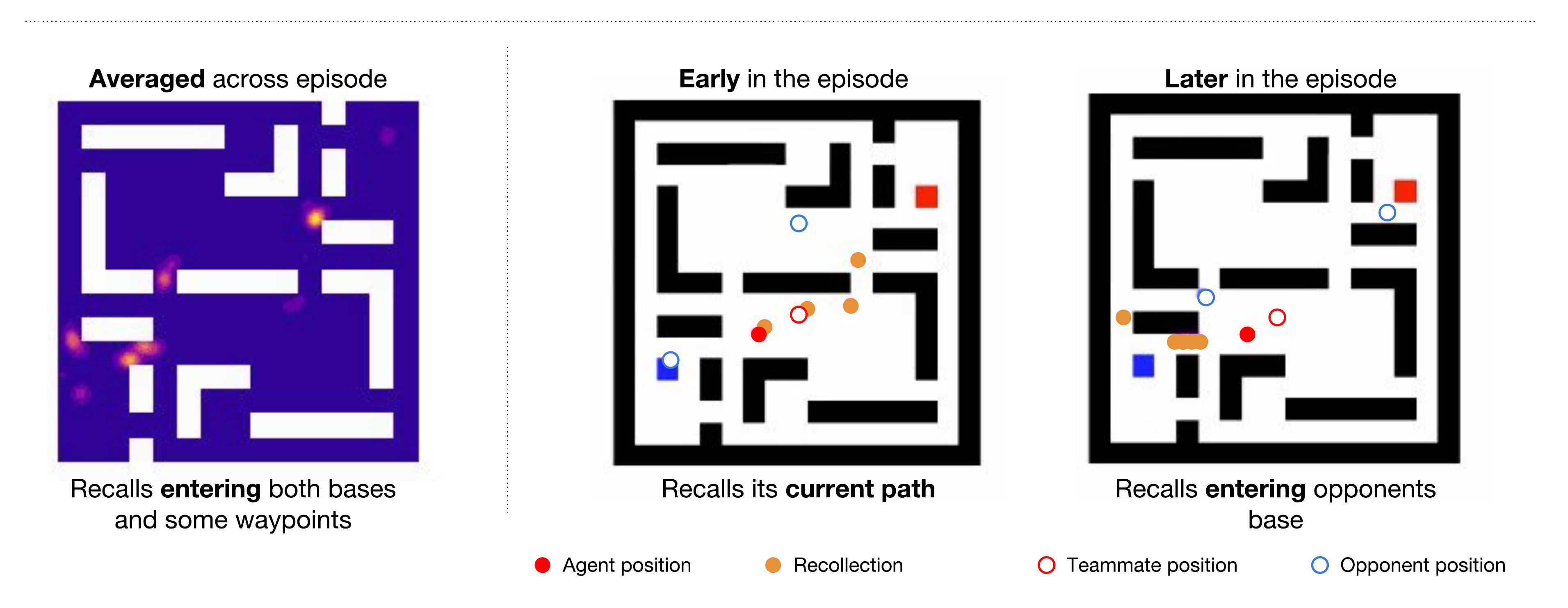}
    \caption{\small {\bf Top:}  Shown are Hinton diagrams representing how often the FTW agent reads memory slots written to at different locations, which are represented in terms of distance to home and opponent base, on 1000 procedurally generated maps, at different points during training. The size of each square represents the difference between the probability of reading from the given location compared to randomly reading from one of the locations visited earlier in the episode.
    Red indicates that the agent reads from this position more often than random, and blue less. At 450K the agent appears to have learned to read from near its own base and just outside the opponent base. {\bf Bottom:} Shown are memory recall patterns for an example episode. The heatmap plot on the left shows memory recall frequency averaged across the episode. Shown on the right are the recall patterns during the agent's first exploration of a newly encountered map. Early in the episode, the agent simply recalls its own path. In almost the same situation later in the episode, the agent recalls entering the opponent base instead.
    }
    \label{fig:ext_dnc}
\end{figure}

\begin{figure}[t]
    \centering
    \includegraphics[width=\textwidth]{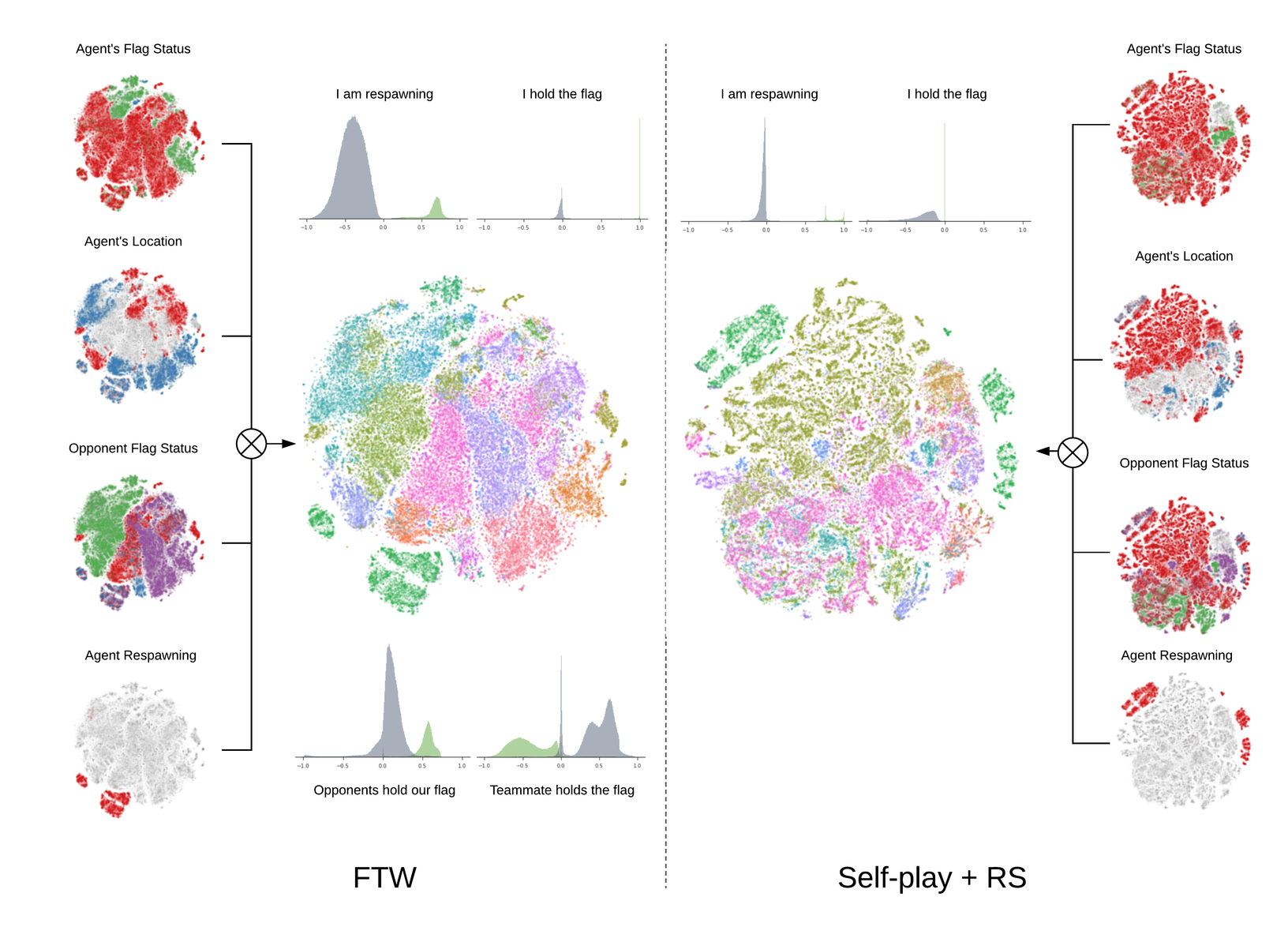}
    \caption{\small Shown is a side-by-side comparison of the internal representations learned from playing CTF for the FTW and Self-play + RS agent, visualised using t-SNE and single neuron activations (Figure~\ref{fig:three} for more information). The self-play agent's representation is seen to be significantly less coherently clustered by game state, especially with respect to flag possessions. Furthermore, it appears to have developed only two highly selective neurons compared to four for the FTW agent.}
    \label{fig:ext_tsnes}
\end{figure}

\subsection{Agent Behaviour}
\begin{figure}[t]
    \centering
    \includegraphics[width=\textwidth]{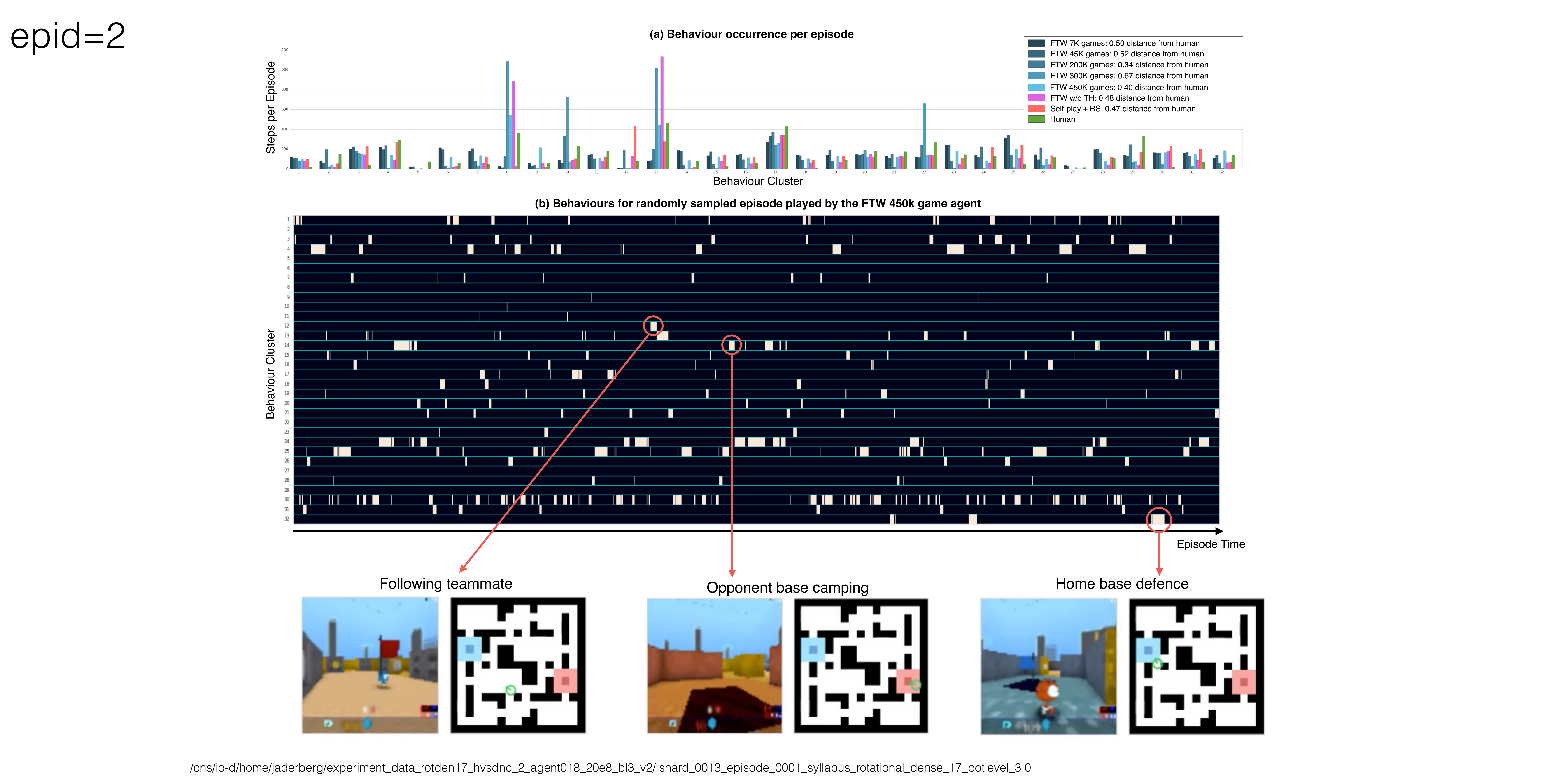}
    \caption{\small {\bf (a)} Shown is a collection of bar plots, one for each of 32 automatically discovered behaviour clusters, representing the number of frames per episode during which the behaviour has been observed  for the FTW agent at different points in training, the FTW agent without the temporal hierarchy (TH), the Self-play + RS agent, and human players, averaged over maps and episodes. The behavioural fingerprint changes significantly throughout training, and differs considerably between models with and without temporal hierarchy. {\bf (b)} Shown is the multi-variate time series of active behaviour clusters during an example episode played by the trained FTW agent. Shown are three particular situations represented  by the behaviour clusters: {\it following your teammate}, {\it enemy base camping}, and {\it home base defence}.}
    \label{fig:ext_behvaiours}
\end{figure}

The CTF games our agents played were five minutes long and consisted of 4500 elemental actions by each player.
To better understand and interpret the behaviour of agents we considered modelling temporal chunks of high-level game features.
We segmented games into two-second periods represented by a sequence of game features (\eg~distance from bases, agent's room, visibility of teammates and opponents, flag status, see Section~\ref{sec:behanalysis}) and used a variational autoencoder (VAE) consisting of an RNN encoder and decoder~\cite{bowman2015generating} to find a compressed vector representation of these two seconds of high-level agent-centric CTF game-play.
We used a Gaussian mixture model (GMM) with 32 components to find clusters of behaviour in the VAE-induced vector representation of game-play segments (Section~\ref{sec:behanalysis} for more details). These discrete cluster assignments allowed us to represent high-level agent play as a sequence of clusters indices (Figure \ref{fig:ext_behvaiours} (b)). These two second behaviour prototypes were interpretable and represented a wide range of meaningful behaviours such as home base camping, opponents base camping, defensive behaviour, teammate following, respawning, and empty room navigation. Based on this representation, high-level agent behaviour could be represented by histograms of frequencies of behaviour prototypes over thousands of episodes. These behavioural fingerprints were shown to vary throughout training, differed strongly between hierarchical and non-hierarchical agent architectures, and were computed for human players as well (Figure \ref{fig:ext_behvaiours} (a)). Comparing these behaviour fingerprints using the Hellinger distance~\cite{hellinger1909neue} we found that the human behaviour was most similar to the FTW agent after 200K games of training.

\section{Experiment Details}
\subsection{Elo Calculation}
\label{sec:supp_elo}

We describe the performance of both agents (human or otherwise) in terms of Elo ratings\cite{ELO}, as commonly used in both traditional games like chess and in competitive video game ranking and matchmaking services. While Elo ratings as described for chess address the one-versus-one case, we extend this for CTF to the $n$-versus-$n$ case by making the assumption that the rating of a team can be decomposed as the sum of skills of its team members.

Given a population of $M$ agents, let $\psi_i \in \mathbb{R}$ be the rating for agent $i$. We describe a given match between two teams, blue and red, with a vector $\vec{m} \in \mathbb{Z}^M$, where $m_i$ is the number of times agent $i$ appears in the blue team less the number of times the agent appears in the red team. Using our additive assumption we can then express the standard Elo formula as:
\begin{equation}
  \mathbb{P}(\text{blue wins against red} | \vec{m}, \vec{\phi}) = \frac{1}{1 + 10^{-\vec{\psi}^T\vec{m}/400}}.
\end{equation}

To calculate ratings given a set of matches with team assignments $\vec{m}_i$ and outcomes $y_i$ ($y_i=1$ for ``blue beats red'' and $y_i=\frac{1}{2}$ for draw), we optimise $\vec{\psi}$ to find ratings $\vec{\psi}^\ast$ that maximise the likelihood of the data. Since win probabilities are determined only by absolute differences in ratings we typically anchor a particular agent (Bot 4) to a rating of 1000 for ease of interpretation.

For the purposes of PBT, we calculate the winning probability of $\pi_i$ versus $\pi_j$ using $\vec{m}_i = 2$ and $\vec{m}_j = -2$ (and $\vec{m}_k=0$ for $k \notin \{i,j\}$), \ie~we assume that both players on the blue team are $\pi_i$ and similarly for the red team.

\subsection{Environment Observation and Action Space}
\label{sec:supp_env_spec}

DeepMind Lab\cite{beattie2016deepmind} is capable of rendering colour observations at a wide range of resolutions. We elected to use a resolution of 84$\times$84 pixels as in previous related work in this environment\cite{A3C,JaderbergUnreal}. Each pixel is represented by a triple of three bytes, which we scale by $\tfrac{1}{255}$ to produce an observation $\vx_t \in [0, 1]^{84\times84\times3}$.

The environment accepts actions as a composite of six types of partial actions: change in yaw (continuous), change in pitch (continuous), strafing left or right (ternary), moving forward or backwards (ternary), tagging and jumping (both binary). To further simplify this space, we expose only two possible values for yaw rotations (10 and 60) and just one for pitch (5). Consequently, the number of possible composite actions that the agent can produce is of size $5 \cdot 3 \cdot 3 \cdot 3 \cdot 2 \cdot 2 = 540$.

\subsection{Procedural Environments}\label{sec:procmap}

\paragraph{Indoor Procedural Maps}
The procedural indoor maps are flat, point-symmetric mazes consisiting of rooms connected by corridors.
Each map has two base rooms which contain the team's flag spawn point and several possible player spawn points.
Maps are contextually coloured: the red base is red, the blue base is blue, empty rooms are grey and narrow corridors are yellow.
Artwork is randomly scattered around the map's walls.

The procedure for generating an indoor map is as follows:

\begin{enumerate}
    \item Generate random sized rectangular rooms within a fixed size square area (\eg~$13 \times 13$ or $17 \times 17$ cells). Room edges were only placed on even cells meaning rooms always have odd sized walls. This restriction was used to work with the maze backtracking algorithm.
    \item Fill the space between rooms using the backtracking maze algorithm to produce corridors. Backtracking only occurs on even cells to allow whole cell gaps as walls.
    \item Remove dead ends and horseshoes in the maze.
    \item Searching from the top left cell, the first room encountered is declared the base room. This ensures that base rooms are typically at opposite ends of the arena.
    \item The map is then made to be point-symmetric by taking the first half of the map and concatenating it with its reversed self.
    \item Flag bases and spawn points are added point-symmetrically to the base rooms.
    \item The map is then checked for being solveable and for meeting certain constraints (base room is at least 9 units in area, the flags are a minimum distance apart).
    \item Finally, the map is randomly rotated (to prevent agents from exploiting the skybox for navigation).
\end{enumerate}

\paragraph{Outdoor Procedural Maps}
The procedural outdoor maps are open and hilly naturalistic maps containing obstacles and rugged terrain. Each team's flag and spawn locations are on opposite corners of the map. Cacti and boulders of random shapes and sizes are scattered over the landscape. To produce the levels, first the height map was generated using the diamond square fractal algorithm. This algorithm was run twice, first with a low variance and then with a high variance and compiled using the element-wise max operator. Cacti and shrubs were placed in the environment using rejection sampling. Each plant species has a preference for a distribution over the height above the water table. After initial placement, a lifecycle of the plants was simulated with seeds being dispersed near plants and competition limiting growth in high-vegetation areas. Rocks were placed randomly and simulated sliding down terrain to their final resting places. After all entities had been placed on the map, we performed pruning to ensure props were not overlapping too much. Flags and spawn points were placed in opposite quadrants of the map. The parameters of each map (such as water table height, cacti, shrub and rock density) were also randomly sampled over each individual map. 1000 maps were generated and 10 were reserved for evaluation.

\subsection{Training Details}
\label{sec:supp_train}

Agents received observations from the environment 15 times (steps) per second. For each observation, the agent returns an action to the environment, which is repeated four times within the environment\cite{A3C,JaderbergUnreal}. Every training game lasts for five minutes, or, equivalently, for 4500 agent steps. Agents were trained for two billion steps, corresponding to approximately 450K games.

Agents' parameters were updated every time a batch of 32 trajectories of length 100 had been accumulated from the arenas in which the respective agents were playing. We used RMSProp\cite{RMSPROP} as the optimiser, with epsilon $10^{-5}$, momentum $0$, and decay rate $0.99$. The initial learning rate was sampled per agent from LogUniform$(10^{-5},5\cdot 10^{-3})$ and further tuned during training by PBT, with a population size of 30. Both V-Trace clipping thresholds $\bar \rho, \bar c$ were set to 1. RL discounting factor $\gamma$ was set to 0.99.

All agents were trained with at least the components of the UNREAL loss\cite{JaderbergUnreal}: the losses used by A3C\cite{A3C}, plus pixel control and reward prediction auxiliary task losses. The baseline cost weight was fixed at $0.5$, the initial entropy cost was sampled per agent from LogUniform$(5 \cdot 10^{-4},10^{-2})$, the initial reward prediction loss weight was sampled from LogUniform$(0.1,1)$, and the initial pixel control loss weight was sampled from LogUniform$(0.01,0.1)$. All weights except the baseline cost weight were tuned during training by PBT.

Due to the composite nature of action space, instead of training pixel control policies directly on 540 actions, we trained independent pixel control policies for each of the six action groups.

The reward prediction loss was trained using a small replay buffer, as in UNREAL\cite{JaderbergUnreal}. In particular, the replay buffer had capacity for 800 non-zero-reward and 800 zero-reward sequences. Sequences consisted of three observations. The batch size for the reward prediction loss was 32, the same as the batch size for all the other losses. The batch consisted of 16 non-zero-reward sequences and 16 zero-reward sequences.

For the FTW agent with temporal hierarchy, the loss includes the KL divergence between the prior distribution (from the slow-ticking core) and the posterior distribution (from the fast-ticking core), as well as KL divergence between the prior distribution and a multivariate Gaussian with mean 0, standard deviation 0.1.
The weight on the first divergence was sampled from $\text{LogUniform}(10^{-3},1)$,
and the weight on the second divergence was sampled from $\text{LogUniform}(10^{-4},10^{-1})$.
A scaling factor on the gradients flowing from fast to slow ticking cores was sampled from $\text{LogUniform}(0.1,1)$.
Finally, the initial slower-ticking core time period $\tau$ was sampled from Categorical$([5,6,\dots, 20])$.
These four quantities were further optimised during training by PBT.

\subsubsection{Training Games}\label{sec:traingames}

Each training CTF game was started by randomly sampling the level to play on. For indoor procedural maps, first (with 50\% probability) the size of map (13 or 17) and its geometry were generated according to the procedure described in Section~\ref{sec:procmap}. For outdoor procedural maps one of the 1000 pre-generated maps was sampled uniformly.
Next, a single agent $\pi_p$ was randomly sampled from the population. Based on its Elo score three more agents were sampled without replacement from the population according to the distribution
$$
\forall_{\pi \in \vec{\pi}_{-p}} \mathbb{P}(\pi|\pi_p) \propto \tfrac{1}{\sqrt{2\pi \sigma^2}} \exp\left ( - \frac{
(\mathbb{P}({\pi_p\text{  beats }\pi} | \phi) - 0.5)^2
}{2\sigma^2}  \right )\;\;\;\;\;\text{ where } \sigma = \tfrac{1}{6}
$$
which is a normal distribution over Elo-based probabilities of winning, centred on agents of the same skill.
For self-play ablation studies agents were paired with their own policy instead.
The agents in the game pool were randomly assigned to the red and blue teams.
After each 5 minute episode this process was repeated.

\begin{figure}
    \centering
    \includegraphics[width=\textwidth]{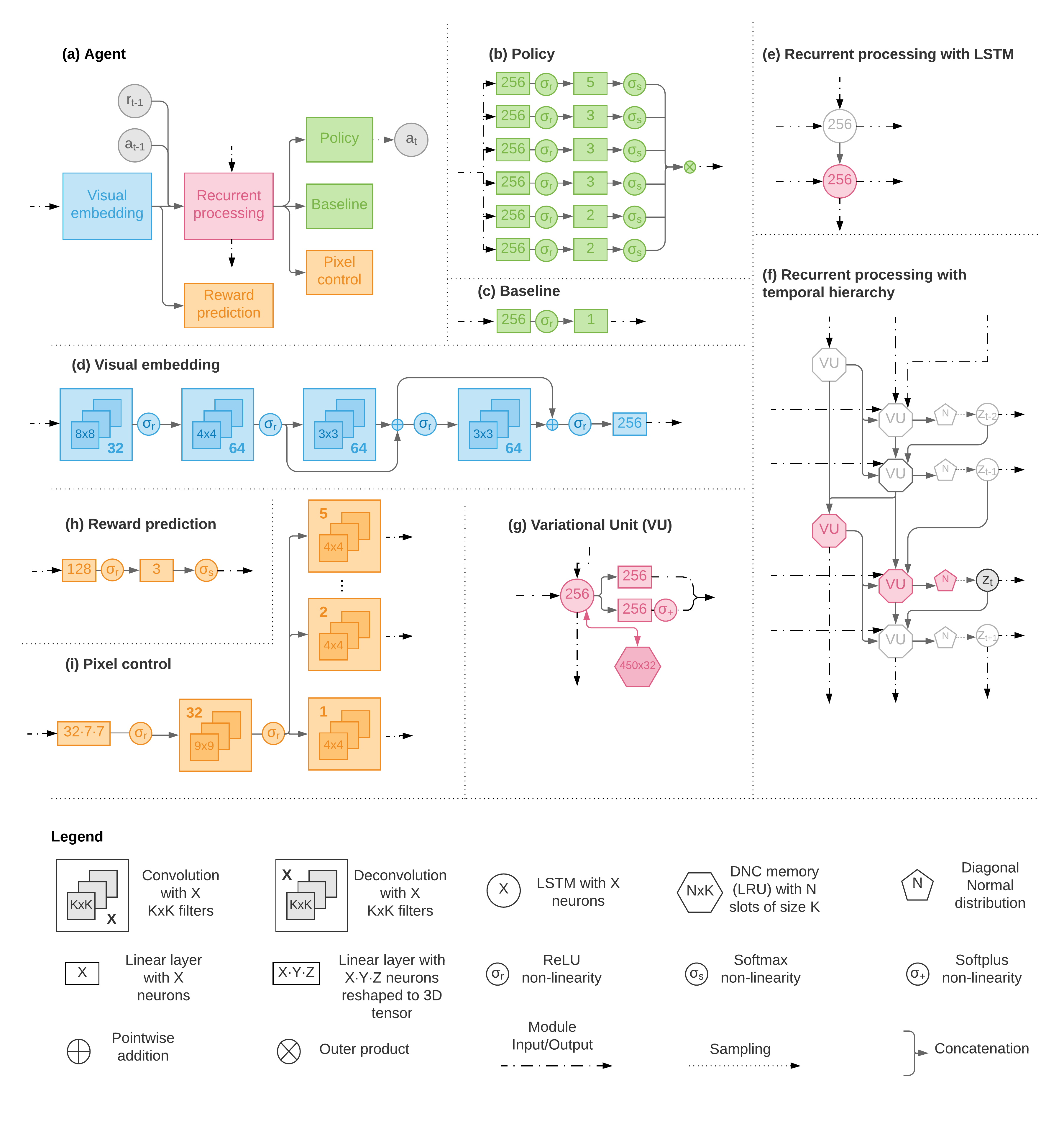}
    \caption{\small Shown are network architectures of agents used in this study. All agents have the same high-level architecture {\bf (a)}, using a decomposed policy {\bf (b)} (see Section~\ref{sec:supp_env_spec}), value function {\bf (c)}, and convolutional neural network (CNN) visual feature extractor {\bf (d)}. The baseline agents and ablated FTW without temporal hierarchy agents use an LSTM for recurrent processing {\bf (e)}. The FTW agent uses a temporal hierarchy for recurrent processing {\bf (f)} which is composed of two variational units {\bf (g)}. All agents use reward prediction {\bf (h)} and independent pixel control {\bf (i)} auxiliary task networks.}
    \label{fig:arch}
\end{figure}

\subsection{Game Events}\label{sec:events}
There are 13 binary game events with unique game point values $\rho_t$. These events are listed below, along with the default values $\vec{w}_\text{quake}$ from the default Quake III Arena points system used for manual reward shaping baselines (Self-play + RS, PBT + RS):
\begin{equation*}
\begin{aligned}
\rho^{(1)}_t &= \text{I am tagged with the flag}    & & p^{(1)} = -1 & & \vec{w}^{(1)}_\text{quake} = 0 \\
\rho^{(2)}_t &= \text{I am tagged without the flag} & & p^{(2)} = -1 & & \vec{w}^{(2)}_\text{quake} = 0 \\
\rho^{(3)}_t &= \text{I captured the flag}          & & p^{(3)} = 1  & & \vec{w}^{(3)}_\text{quake} = 6 \\
\rho^{(4)}_t &= \text{I picked up the flag}         & & p^{(4)} = 1 & & \vec{w}^{(4)}_\text{quake} = 1 \\
\rho^{(5)}_t &= \text{I returned the flag}          & & p^{(5)} = 1 & & \vec{w}^{(5)}_\text{quake} = 1 \\
\rho^{(6)}_t &= \text{Teammate captured the flag}   & & p^{(6)} = 1 & & \vec{w}^{(6)}_\text{quake} = 5 \\
\rho^{(7)}_t &= \text{Teammate picked up the flag}  & & p^{(7)} = 1 & & \vec{w}^{(7)}_\text{quake} = 0 \\
\rho^{(8)}_t &= \text{Teammate returned the flag}   & & p^{(8)} = 1 & & \vec{w}^{(8)}_\text{quake} = 0 \\
\rho^{(9)}_t &= \text{I tagged opponent with the flag} & &  p^{(9)} = 1 & & \vec{w}^{(9)}_\text{quake} = 2 \\
\rho^{(10)}_t &= \text{I tagged opponent without the flag} & & p^{(10)} = 1 & & \vec{w}^{(10)}_\text{quake} = 1 \\
\rho^{(11)}_t &= \text{Opponents captured the flag} & & p^{(11)} = -1  & & \vec{w}^{(11)}_\text{quake} = 0 \\
\rho^{(12)}_t &= \text{Opponents picked up the flag} & & p^{(12)} = -1 & & \vec{w}^{(12)}_\text{quake} = 0 \\
\rho^{(13)}_t &= \text{Opponents returned the flag} & & p^{(13)} = -1  & & \vec{w}^{(13)}_\text{quake} = 0
\end{aligned}
\end{equation*}
Agents did not have direct access to these events.
FTW agents' initial internal reward mapping was sampled independently for each agent in the population according to
$$\vec{w}^{(i)}_\text{initial} = \epsilon \cdot p^{(i)}\;\;\;\;\;\;\;\; \epsilon \sim \text{LogUniform}(0.1, 10.0).$$
after which it was adapted through training with reward evolution.

\subsection{Ablation}\label{sec:ablation}
We performed two separate series of ablation studies, one on procedural indoor maps and one on procedural outdoor maps. For each environment type we ran the following experiments:
\begin{itemize}
\item {\bf Self-play:} An agent with an LSTM recurrent processing core (Figure~\ref{fig:arch} (e)) trained with the UNREAL loss functions described in Section~\ref{sec:supp_train}.
Four identical agent policies played in each game, two versus two.
Since there was only one agent policy trained, no Elo scores could be calculated, and population-based training was disabled.
A single reward was provided to the agent at the end of each episode, +1 for winning, -1 for losing and 0 for draw.
\item {\bf Self-play + Reward Shaping:} Same setup as Self-play above, but with manual reward shaping given by $\vec{w}_\text{quake}$.
\item {\bf PBT + Reward Shaping}: Same agent and losses as Self-play + Reward Shaping above, but for each game in each arena the four participating agents were sampled without replacement from the population using the process described in Section~\ref{sec:supp_train}. Based on the match outcomes Elo scores were calculated for the agents in the population as described in Section~\ref{sec:supp_elo}, and were used for PBT.
\item {\bf FTW w/o Temporal Hierarchy}: Same setup as PBT + Reward Shaping above, but with Reward Shaping replaced by an internal reward signals evolved by PBT.
\item {\bf FTW}: The FTW agent, using the recurrent processing core with temporal hierarchy (Figure~\ref{fig:arch} (f)), with the training setup described in Methods: matchmaking, PBT, and internal reward signal.
\end{itemize}

\subsection{Distinctly Selective Neurons}
For identifying the neuron in a given agent that is most selective for a game state feature $y$ we recorded 100 episodes of the agent playing against Bot 3. Given this dataset of activations $\vec{h}_i$ and corresponding labels $y_i$ we fit a Decision Tree of depth 1 using Gini impurity criterion. The decision tree learner selects the most discriminative dimension of $\vec{h}$ and hence the neuron most selective for $y$. If the accuracy of the resulting stump exceeds 97\%  over $100 \cdot 4500$ steps we consider it to be a \emph{distinctly selective neuron}.

\subsection{Behavioural Analysis}\label{sec:behanalysis}
For the behavioural analysis, we model chunks of two seconds (30 agent steps) of gameplay. Each step is represented by 56 agent-centric binary features derived from groundtruth game state:
\begin{itemize}
 \item (3 features) Thresholded shortest path distance from other three agents.
 \item (4 features) Thresholded shortest path distance from each team's base and flags.
 \item (4 features) Whether an opponent captured, dropped, picked up, or returned a flag.
 \item (4 features) Whether the agent captured, dropped, picked up, or returned a flag.
 \item (4 features) Whether the agent's teammate captured, dropped, picked up, or returned a flag.
 \item (4 features) Whether the agent was tagged without respawning, was tagged and must respawn, tagged an opponent without them respawning, or tagged an opponent and they must respawn.
 \item (4 features) What room an agent is in: home base, opponent base, corridor, empty room.
 \item (5 features) Visibility of teammate (visible and not visible), no opponents visible, one opponent visible, two opponents visible.
 \item (5 features) Which other agents are in the same room: teammate in room, teammate not in room, no opponents in room, one opponent in room, two opponents in room.
 \item (4 features) Each team's base visibility.
 \item (13 features) Each team's flag status and visibility. Flags status can be either at base, held by teammate, held by opponent, held by the agent, or stray.
 \item (2 features) Whether agent is respawning and cannot move or not.
\end{itemize}

For each of the agents analysed, 1000 episodes of pairs of the agent playing against pairs of Bot 3 were recorded and combined into a single dataset. A variational autoencoder (VAE)\cite{rezende2014stochastic, kingma2013auto} was trained on batches of this mixed agent dataset (each data point has dimensions 30$\times$56) using an LSTM encoder (256 units) over the 30 time steps, whose final output vector is linearly projected to a 128 dimensional latent variable (diagonal Gaussian). The decoder was an LSTM (256 units) which took in the sampled latent variable at every step.

After training the VAE, a dataset of 400K data points was sampled, the latent variable means computed, and a Gaussian mixture model (GMM) was fit to this dataset of 400K$\times$128, with diagonal covariance and 32 mixture components. The resulting components were treated as behavioural clusters, letting us characterise a two second clip of CTF gameplay as one belonging to one of 32 behavioural clusters.

\subsection{Bot Details}\label{sec:botdetails}
The bots we use for evaluation are a pair of Tauri and Centuri bots from Quake III Arena as defined below.

\begin{center}
\begin{tabular}{  l| c c c c c | c c c c c }
\small
  Bot Personality & \multicolumn{5}{c|}{Tauri} & \multicolumn{5}{c}{Centauri} \\
  \hline
  Bot Level & 1 & 2 & 3 & 4 & 5       & 1 & 2 & 3 & 4 & 5  \\
  \hline
  ATTACK\_SKILL & 0.0 & 0.25 & 0.5 & 1.0 & 1.0    & 0.0 & 0.25 & 0.5 & 1.0 & 1.0  \\
  AIM\_SKILL & 0.0 & 0.25 & 0.5 & 1.0 & 1.0       & 0.0 & 0.25 & 0.5 & 1.0 & 1.0  \\
  AIM\_ACCURACY & 0.0 & 0.25 & 0.5 & 1.0 & 1.0    & 0.0 & 0.25 & 0.5 & 1.0 & 1.0  \\
  VIEW\_FACTOR & 0.1 & 0.35 & 0.6 & 0.9 & 1.0   & 0.1 & 0.35 & 0.6 & 0.9 & 1.0  \\
  VIEW\_MAXCHANGE & 5 & 90 & 120 & 240 & 360    & 5 & 90 & 120 & 240 & 360  \\
  REACTIONTIME & 5.0 & 4.0 & 3.0 & 1.75 & 0.0   & 5.0 & 4.0 & 3.0 & 1.75 & 0.0  \\
  \hline
  CROUCHER & 0.4 & 0.25 & 0.1 & 0.1 & 0.0       & 0.4 & 0.25 & 0.1 & 0.1 & 0.0 \\
  JUMPER & 0.4 & 0.45 & 0.5 & 1.0 & 1.0         & 0.4 & 0.45 & 0.5 & 1.0 & 1.0  \\
  WALKER & 0.1 & 0.05 & 0.0 & 0.0 & 0.0         & 0.1 & 0.05 & 0.0 & 0.0 & 0.0  \\
  WEAPONJUMPING & 0.1 & 0.3 & 0.5 & 1.0 & 1.0   & 0.1 & 0.3 & 0.5 & 1.0 & 1.0  \\
  GRAPPLE\_USER & 0.1 & 0.3 & 0.5 & 1.0 & 1.0   & 0.1 & 0.3 & 0.5 & 1.0 & 1.0 \\
  AGGRESSION &0.1 & 0.3 & 0.5 & 1.0 & 1.0       &0.1 & 0.3 & 0.5 & 1.0 & 1.0 \\
  SELFPRESERVATION & 0.1 & 0.3 & 0.5 & 1.0 & 1.0& 0.1 & 0.3 & 0.5 & 1.0 & 1.0 \\
  VENGEFULNESS & 0.1 & 0.3 & 0.5 & 1.0 & 1.0    & 0.1 & 0.3 & 0.5 & 1.0 & 1.0 \\
  CAMPER & 0.0 & 0.25 & 0.5 & 0.5 & 0.0         & 0.0 & 0.25 & 0.5 & 0.5 & 0.0 \\
  EASY\_FRAGGER & 0.1 & 0.3 & 0.5 & 1.0 & 1.0   & 0.1 & 0.3 & 0.5 & 1.0 & 1.0 \\
  ALERTNESS & 0.1 & 0.3 & 0.5 & 1.0 & 1.0       & 0.1 & 0.3 & 0.5 & 1.0 & 1.0 \\
  \hline
  AIM\_ACCURACY & 0.0 & 0.22 & 0.45 & \textbf{0.75} & 1.0 & 0.0 & 0.22 & 0.45 & \textbf{0.95} & 1.0 \\
  FIRETHROTTLE & 0.01 & 0.13 & 0.25 & \textbf{1.0} & \textbf{1.0} & 0.01 & 0.13 & 0.25 & \textbf{0.1} & \textbf{0.01} \\
\end{tabular}
\end{center}

\end{document}